\pdfoutput=1
\documentclass[11pt,letterpaper]{article}

\usepackage[T1]{fontenc}
\usepackage[letterpaper,margin=1in]{geometry}
\usepackage{amsmath,amssymb,amsthm,mathtools}
\usepackage{thm-restate}
\usepackage{graphicx}
\usepackage{needspace}
\usepackage{enumitem}
\usepackage{algorithm}
\floatstyle{ruled}
\newfloat{construction}{htbp}{loc}
\floatname{construction}{Construction}
\usepackage{algpseudocode}
\algrenewcommand\alglinenumber[1]{\scriptsize #1}
\usepackage{booktabs}
\usepackage{microtype}
\usepackage{url}
\usepackage[normalem]{ulem}
\usepackage[dvipsnames]{xcolor}
\definecolor{deeprednew}{rgb}{0.58,0.0,0.0}
\definecolor{darkgreennew}{rgb}{0.0,0.31,0.11}
\usepackage[backend=bibtex,style=alphabetic,natbib=true,sortcites=true,maxcitenames=5,maxbibnames=99,maxalphanames=6,minalphanames=3,isbn=false,doi=false,url=false,backref=true]{biblatex}
\DeclareFieldFormat[misc]{title}{\iffieldundef{eprint}{\mkbibemph{#1}}{\mkbibquote{#1\isdot}}}
\usepackage[colorlinks=true,linkcolor=deeprednew,
            citecolor=darkgreennew,urlcolor=blue]{hyperref}
\usepackage[capitalise,nameinlink]{cleveref}
\makeatletter
\providecommand*{\theHALG@line}{}
\renewcommand*{\theHALG@line}{\thealgorithm.\arabic{ALG@line}}
\makeatother
\crefname{lemma}{Lemma}{Lemmas}
\crefname{fact}{Fact}{Facts}
\crefname{claim}{Claim}{Claims}
\crefname{theorem}{Theorem}{Theorems}
\crefname{proposition}{Proposition}{Propositions}
\crefname{corollary}{Corollary}{Corollaries}
\crefname{observation}{Observation}{Observations}
\crefname{remark}{Remark}{Remarks}
\crefname{conjecture}{Conjecture}{Conjectures}
\crefname{definition}{Definition}{Definitions}
\crefname{problem}{Problem}{Problems}
\crefname{subsection}{Subsection}{Subsections}
\crefname{algorithm}{Algorithm}{Algorithms}
\crefname{construction}{Construction}{Constructions}
\usepackage{mathpazo}
\usepackage{pgfplots}
\pgfplotsset{compat=1.18}
\usetikzlibrary{arrows.meta,calc,positioning,patterns}
\colorlet{boundupper}{darkgreennew}
\colorlet{boundlower}{deeprednew}

\allowdisplaybreaks
\setlist{nosep,leftmargin=*}

\newtheorem{theorem}{Theorem}[section]
\newtheorem{lemma}[theorem]{Lemma}
\newtheorem{fact}[theorem]{Fact}
\newtheorem{proposition}[theorem]{Proposition}
\newtheorem{corollary}[theorem]{Corollary}

\theoremstyle{definition}
\newtheorem{claim}[theorem]{Claim}

\newcommand{\E}{\mathbb{E}}
\newcommand{\Prb}{\mathbb{P}}
\DeclareMathOperator{\Var}{Var}

\DeclareMathOperator{\TV}{TV}
\DeclareMathOperator{\KL}{KL}
\DeclareMathOperator{\supp}{supp}
\DeclareMathOperator{\tr}{tr}
\DeclareMathOperator{\sign}{sign}
\DeclareMathOperator*{\argmin}{arg\,min}
\newcommand{\Amp}{\mathop{\mathsf{Amp}}\nolimits}
\newcommand{\MW}{\mathop{\mathsf{MW}}\nolimits}
\newcommand{\Lap}{\mathop{\mathsf{Lap}}\nolimits}
\newcommand{\1}{\mathbf{1}}
\newcommand{\N}{\mathbb{N}}

\newcommand{\mper}{\, .}

\newcommand{\Xcal}{\mathcal X}
\newcommand{\dde}{\mathcal G}
\newcommand{\phat}{\hat p}
\usepackage{authblk}
\newif\iffulldraft
\fulldrafttrue

\title{Data-Dependent Probability Estimation\\ via Sample Amplification}
\author[1]{Vitaly Feldman}
\author[1]{Satyen Kale}
\author[2]{Mahbod Majid\thanks{Work done when the author was an intern at Apple}}
\author[1]{Kunal Talwar}
\affil[1]{Apple}
\affil[2]{MIT}
\hypersetup{pdftitle={Data-Dependent Probability Estimation via Sample Amplification},
            pdfauthor={Vitaly Feldman, Satyen Kale, Mahbod Majid, Kunal Talwar}}

\begin{document}

\iffulldraft
\hypersetup{pageanchor=false}
\pagestyle{empty}
\maketitle
\begin{abstract}
Can we estimate the probability of an event chosen after seeing the data,
without using any fresh samples? Its empirical probability can be arbitrarily
misleading, while learning the entire distribution to constant accuracy
requires $\Theta(d)$ samples on a domain of size $d$.
We show that $\widetilde O(\sqrt d)$
samples suffice for constant mean squared error, provided the rule choosing
the event is known. More generally, our estimator handles arbitrary
$[0,1]$-valued queries and runs in polynomial time given evaluation access
to the rule. This sample complexity matches the known lower bound up to
logarithmic factors and resolves the open question posed by Feldman, Kale,
Raman, Talwar, and Tewari~\cite{FeldmanKRTT26}.

Our estimation framework relies on a new combination of sample amplification~\citep{axelrod2020amplification,axelrod2024statistical} with
distribution learning via online optimization.
Our estimator achieves mean squared error $O(d^{3/4}/n)$ when
$d\leq n$ and $\widetilde O(d^{1/4}/\sqrt n)$ when $d>n$. We complement this with a lower bound of $\Omega(\min\{1,\sqrt d/n\})$ on the
mean squared error. Using our framework, we also design an efficient estimator for data-dependent queries on Gaussian distributions with unknown mean and identity covariance. Finally, we show that a smoothed variant of the problem is much
(exponentially) easier.
For discrete distributions, if the sample size is chosen uniformly from
$\{n,\ldots,2n-1\}$, then $n=O(\log d)$ suffices for constant mean squared
error.
\end{abstract}

\thispagestyle{empty}
\clearpage
\pagestyle{plain}
\pagenumbering{arabic}
\hypersetup{pageanchor=true}
\else
\pagestyle{plain}
\pagenumbering{arabic}
\fi

\section{Introduction}
\label{sec:upper-bounds}

Many core problems in machine learning and statistics ask for the
probability of an event that was chosen using the data. The test error of
a trained classifier is the probability of its error region, which is
determined by the training sample. The missing mass is the probability of
the elements that do not appear in the sample. The data-dependent targets
of adaptive data analysis and post-selection inference have the same
form~\cite{FeldmanKRTT26}. In each case the event depends on the sample,
and so its empirical probability can be far from its true probability. A
classifier that memorizes its training examples has zero training error
even if it errs on most of the distribution, and the empirical missing mass
is always zero. Can we nevertheless estimate the probability of such an
event using only the observations that were used to choose it?

\citet{gastpar2024fantastic,gastpar2024which} study this question for the
test error of a known learning algorithm, as a way to understand when
generalization bounds can be tight. \citet{FeldmanKRTT26} formulate it for
arbitrary events as the problem of data-dependent probability estimation.
In both formulations the rule that chooses the event is known, as it is
when an analyst chooses the event. The estimator can therefore evaluate the
rule on other samples, but it receives no observations beyond those used
to choose the event.

Many estimators have been developed for specific questions of this kind.
The Good--Turing estimator and its variants estimate the missing
mass~\cite{Good1953,mcallester2000convergence,mcallester2003concentration},
and holdout sets, cross-validation~\cite{stone1974cross} and the
bootstrap~\cite{efron1983estimating} estimate the error of a learning
algorithm. However, these estimators either apply to specific problems or
rely on the stability of the rule. A holdout set estimates the probability
of an event chosen from fewer observations, and cross-validation averages
over events chosen from different samples. Without stability, removing
even one observation can change the event completely. Thus these other
events may say little about the one we want to
estimate~\cite{kearns1999algorithmic}.

For an arbitrary rule, the only general solution that we are aware of is
to learn the distribution well enough to answer every bounded query. On a
domain of size $d$, this requires $\Theta(d)$ samples for constant
accuracy (see, e.g., \citep{Canonne2020shortnote}). At the same time, known constructions show that $\Omega(\sqrt d)$
samples are necessary~\cite{gastpar2024which,FeldmanKRTT26}. In terms of
the domain size, some rules on domains of size $O(n^2)$ choose events that
cannot be estimated, while every rule on a domain of size $o(n)$ can be
handled. The remaining gap (posed as an open problem in \cite{FeldmanKRTT26}) boils down to the following question:

\begin{quote}
\centering
\emph{Does data-dependent probability estimation need as many samples\\
as learning the entire distribution?}
\end{quote}

Naturally, it is also useful to understand for which families of distributions (beyond arbitrary ones over a fixed support)
this problem can be solved.%

\Needspace{12\baselineskip}
\subsection{Our Results}

We show that the answer is no by giving an efficient estimator that, for
every rule on a domain of size $d$, achieves any fixed mean squared error
from $\widetilde O(\sqrt d)$ samples. Our solution is based on an
algorithmic framework that combines sample
amplification~\cite{axelrod2020amplification,axelrod2024statistical} with
estimation of the distribution via stochastic optimization using the multiplicative weights method~\citep{arora2012multiplicative}.

To state our results more formally, let $D$ be an unknown distribution on
$[d]=\{1,\ldots,d\}$, and let $S=(X_1,\ldots,X_n)\sim D^n$. A {\em
data-dependent event} is an event $\dde(S)\subseteq[d]$ chosen by a known
rule, given by a mapping $\dde$ from samples to events. The goal is to
estimate its probability $D(\dde(S))$ from $S$. More generally, $\dde(S)$ can be a
query $\dde(S)\colon[d]\to[0,1]$ with value $\E_D[\dde(S)]$ on $D$ (we
identify an event with its indicator function). The estimator knows $\dde$
and can evaluate it on other samples (where one evaluation returns the vector $\dde(s)\in[0,1]^{[d]}$). We measure its error by the mean
squared error over $S$ and the estimator's randomness. We write
$\Delta_d$ for the set of distributions on $[d]$. Our main result for
discrete distributions is the following.

\begin{restatable}[Discrete distributions]{theorem}{mirrordescentupperbound}
\label[theorem]{thm:mirror-descent-upper-bound}
Let $n\geq1$ and $d\geq2$. For every mapping
$\dde\colon[d]^n\to[0,1]^{[d]}$, there is an explicit randomized estimator
$\phat$ such that
\[
\sup_{D\in\Delta_d}
\E_{S\sim D^n}\left[\left(\phat(S)-\E_D[\dde(S)]\right)^2\right]
=O\left(\min\left\{1,\frac{d^{1/4}}{\sqrt n}
\sqrt{\log\left(1+\frac dn\right)}\right\}\right).
\]
The estimator uses at most $\lceil n/\sqrt d\rceil$ evaluations of
$\dde$ and $O((1+n/\sqrt d)(n+d\log^3(nd)))$ time.
\end{restatable}

In particular, the error tends to zero whenever $d=o(n^2/\log^2n)$. For
$n\geq d$, the bound is $O(d^{3/4}/n)$. For comparison, we note that
learning $D$ gives mean squared error $O(d/n)$ for every query. Our bound
gives constant error with $O(\sqrt d\log d)$ samples, instead of the
$\Theta(d)$ samples needed to learn $D$. For $n\geq d$, it improves on
$d/n$ by a factor of $d^{1/4}$. We describe the estimator in
\Cref{sec:overview} and prove \Cref{thm:mirror-descent-upper-bound} in
\Cref{sec:penalized-mw}.

This sample complexity is nearly optimal. The constructions
in~\cite[Theorems~7.2 and~7.3]{FeldmanKRTT26} give a constant lower bound
when $n=O(\sqrt d)$. By extending and strengthening the constructions
in~\citep{FeldmanKRTT26}, we prove the following lower bound.

\begin{restatable}[Lower bound]{theorem}{risklowerbound}
\label[theorem]{thm:risk-lower-bound}
For every $n\geq1$ and $d\geq2$, there is a mapping
$\dde\colon[d]^n\to\{0,1\}^{[d]}$ such that every randomized estimator
$\phat$ satisfies
\[
\sup_{D\in\Delta_d}
\E_{S\sim D^n}\left[\left(\phat(S)-\E_D[\dde(S)]\right)^2\right]
=\Omega\left(\min\left\{1,\frac{\sqrt d}{n}\right\}\right).
\]
\end{restatable}

Thus $\widetilde\Theta(\sqrt d)$ samples are
necessary and sufficient for a sufficiently small constant error. For
$n\geq d$, the bounds proved here leave a polynomial gap between
$\sqrt d/n$ and $d^{3/4}/n$.

Importantly, our guarantee is for the mean squared error. Estimating the
probability of a selected event to accuracy $1/16$ with failure
probability $o(1/\sqrt n)$ requires $\Omega(d)$ samples, as many as
learning the distribution (\Cref{cor:confidence-lower-bound}).

\paragraph{Gaussian distributions.}
As an additional application of our framework we consider the Gaussian
distributions $\mathcal N(\mu,I_d)$ on $\mathbb R^d$ with unknown mean
$\mu$ and identity covariance. We use the amplifier
of~\cite[Examples~4.1 and~4.2]{axelrod2024statistical}
to obtain the following guarantees.

\begin{theorem}[Gaussian distributions, informal]
\label[theorem]{thm:gaussian-informal}
For every mapping that selects a query $f_S\colon\mathbb R^d\to[0,1]$ from
a sample $S\sim\mathcal N(\mu,I_d)^n$, there is an estimator whose mean
squared error is $O(\min\{1,d^{3/4}/n\})$ for every $\mu\in\mathbb R^d$.
For $n\geq d^{5/4}$ and $d\geq\log n$, this bound can be
achieved in polynomial time given evaluation access to
$(s,x)\mapsto f_s(x)$.
\end{theorem}

Thus $O(d^{3/4})$ samples suffice for any fixed error, compared with the
$\Theta(d)$ samples needed to learn the distribution.
In \Cref{sec:gaussian-estimation} we give the formal statements
(\Cref{thm:gaussian-upper-bound,thm:gaussian-sgd}) and their proofs.

\paragraph{Smoothed sample size.}
In practice, an analysis of the data does not depend too strongly on the
exact number of observations. The sample size can then be thought of as
chosen at random from some interval. We model this by choosing the sample
size uniformly from $\{n,\ldots,2n-1\}$, independently of the
observations. For simplicity
we use this interval, although the result holds for any interval whose
length is a constant fraction of $n$ and extends to shorter intervals.

\begin{theorem}[Smoothed sample size]
\label[theorem]{thm:smoothed-intro}
If the sample size is chosen uniformly from $\{n,\ldots,2n-1\}$,
independently of the observations, then for every mapping there is an
efficient estimator with mean squared error
$O(\sqrt{\log(1+d/n)/n})$ for every $D\in\Delta_d$, in expectation over the observations and the
sample size.
\end{theorem}

Thus the dependence on the domain size is only logarithmic. In particular,
$O(\log d)$ samples suffice for constant error.
In \Cref{sec:random-horizon} we prove the bound for intervals of every
length (\Cref{thm:random-horizon-upper-bound}).

\subsection{Related Work}
\label{sec:related-work}

\paragraph{Data-dependent probability estimation.}
\citet{FeldmanKRTT26} formulate
data-dependent probability estimation and study conditions under which
empirical and leave-one-out estimation are accurate. Empirical estimation
requires a generalization guarantee, while their leave-one-out bound
applies to symmetric mappings that are stable under deletion of one
observation. Our estimators work for arbitrary mappings, with bounds
depending on the domain size. We resolve, up to logarithmic factors, their question about the domain size of mappings that cannot be estimated, and prove upper and lower bounds on the error at finer accuracies.

\paragraph{Estimability of learning algorithms.}
\citet{gastpar2024fantastic} study
whether the test error of a learning algorithm can be estimated from its
training sample, and relate this to tight generalization bounds. For a
known learning algorithm, this is an instance of our problem, up to the
choice of error criterion~\cite[Section~3.3]{FeldmanKRTT26}.
Their subsequent work~\cite[Theorem~2.1]{gastpar2024which} constructs
learning rules whose error cannot be estimated to within a sufficiently
small constant when
$n\leq\sqrt d/10$, even when the rule is known. Their hard distributions
are supported on $2d$ labeled points. For domains of this size, our upper
bound matches this sample threshold up to a logarithmic factor.

\paragraph{Adaptive data analysis.}
\citet{DworkFHPRR15} study
how to answer a sequence of adaptively chosen statistical queries
accurately. They use stability to control the generalization error,
a connection developed further in~\cite{BassilyNSSSU16}. These bounds
also apply to empirical estimation when our query mapping satisfies the
required stability conditions. In adaptive data analysis, the analyst
accesses the sample through answers released by a mechanism. Here the
query may depend on the entire sample, and the estimator knows the mapping
that selects it. As discussed
in~\cite{FeldmanKRTT26}, the lower bounds for adaptive data
analysis~\cite{hardt2014preventing,steinke2015interactive} therefore do not
directly imply lower bounds for our problem.

\paragraph{Sample amplification.}
\citet{axelrod2020amplification} study
how to produce a larger sample that is close in distribution to independent
observations. They show that this is possible even when the distribution
cannot be learned accurately.
\citet{axelrod2024statistical}
develop general amplification procedures and connect their guarantees
to comparison of statistical experiments. For discrete distributions,
we follow their resampling approach by copying one sample and shuffling, although our analysis is different and does not require sample splitting.
For Gaussians with known covariance, we use their construction that
preserves the empirical mean and generates new Gaussian
residuals~\cite[Examples~4.1 and~4.2]{axelrod2024statistical}.
Whereas prior work amplifies a dataset once for a single downstream procedure, we bound the one-step $\chi^2$ divergence of adding one observation and compose amplification steps in reverse order with iterative updates.

\paragraph{Multiplicative weights and mirror descent.}
We use the exponentiated-gradient update for square-loss linear prediction~\cite[Figure~2]{kivinen1997exponentiated}, with the standard relative entropy analysis of multiplicative weights~\cite{arora2012multiplicative}.
Our projected updates use the generalized Pythagorean inequality~\cite[Section~2.2]{arora2012multiplicative}, with linear-time KL projection onto distributions whose coordinates are at least $1/(2d)$~\cite[Theorem~7]{HerbsterWarmuth2001}.
The penalized update is an instance of composite objective mirror descent~\cite[Equation~(3)]{COMID2010} with a sample-dependent penalty.
For smoothed sample size (\Cref{sec:random-horizon}), we combine exponentiated-gradient updates using fresh feedback with the relative entropy bound for Laplace prediction~\cite[Proposition~4]{MourtadaGaiffas2022}.
Averaging over the sample size converts the online prediction bound into an estimation guarantee, following the principle of online-to-batch conversion~\cite{CesaBianchiConconiGentile2004}.
\Needspace{20\baselineskip}
\section{Technical Overview}
\label{sec:overview}

We first explain why the natural uses of sample amplification fail in the
discrete setting and how we instead use amplified samples for
multiplicative weights updates. We then describe the projection and the
penalty that lead to \Cref{thm:simple-mw,thm:mirror-descent-upper-bound}, and
the general form of our framework.
Finally, we explain how the same approach applies to Gaussian distributions and why
a smoothed sample size makes the problem easier. An overview of the lower bound constructions is deferred to \Cref{sec:risk-lower-bounds}.
For a sample $s$ we write
$f_s=\dde(s)$, so that our goal is to estimate $\E_D[f_S]$.

Fresh observations from $D$ would make the problem easy, since the average
of $f_S$ over $L$ observations that are independent of $S$ has mean squared
error at most $1/(4L)$. A natural approach is therefore to generate such
observations from $S$ by sample
amplification~\cite{axelrod2020amplification,axelrod2024statistical}. We
amplify a sample $s\in[d]^n$ by appending a copy of a uniformly chosen
entry and uniformly shuffling the resulting $n+1$ entries, and we write
$\Amp(s)=(s',y)$ for the first $n$ entries and the last entry. For
$S\sim D^n$, the distribution of $\Amp(S)$ is within total variation
distance $\sqrt d/n$ of $D^{n+1}$
(\Cref{lem:efficient-amplification}). However, this guarantee holds for
the shuffled pair $(S',Y)$ and not for $(S,Y)$. Thus $Y$ behaves like a
fresh observation only for queries computed from $S'$, and not for $f_S$.
This limitation is not an artifact of the analysis. Every amplified
observation is one of the points of $S$. Thus averaging $f_S$ over amplified
observations gives a weighted version of the empirical estimate. For the
missing mass, $f_S$ is the indicator of the points that do not appear in
$S$. This estimate is then always zero, even when most of the probability
lies outside $S$.

Since amplified observations cannot be used to evaluate $f_S$ directly, we follow the approach of first constructing an estimate $q$ of $D$, and using it to answer with
$\E_q[f_S]$. Note that this approach is implicit in other estimators: for example the empirical estimator and the Laplace predictor.
 When $n=o(d)$, no estimate is accurate for every bounded query, since
this would require $q$ to be close to $D$ in total variation distance.
However, it suffices for $q$ to be accurate on average over the queries
selected on samples from $D^n$. For a fixed $q$, the mean squared error of
$\E_q[f_S]$ is
\[
F(q)=\E_{S\sim D^n}\left[\bigl(\E_q[f_S]-\E_D[f_S]\bigr)^2\right],
\]
which is a convex function of $q\in\Delta_d$. Given a pair of fresh
samples $(S',Y)\sim D^{n+1}$, we can estimate the gradient of $F$ at
$q$ without bias by $2(\E_q[f_{S'}]-f_{S'}(Y))f_{S'}$. These estimates have
coordinates in $[-2,2]$. For optimization over the simplex, this suggests
mirror descent with the entropy regularizer, that is, multiplicative
weights (MW)~\cite{arora2012multiplicative}, updating each coordinate $q(x)$ multiplicatively using the query $f_{S'}$ and observation $Y$.
Importantly, $F$ does not depend on the sample $S$ that we are given.
Therefore its gradient estimates require only pairs that behave like $n+1$
fresh observations, such as the pairs produced by amplification.

\paragraph{Multiplicative weights with amplified samples.}
This suggests an algorithm that starts from an estimate built from $S$,
performs MW updates that each use a new pair $(S',Y)=\Amp(S)$, and answers
the query $f_S$. Unfortunately, the pairs used in different updates are all
generated from $S$ and are strongly dependent. Thus this algorithm is not
quite stochastic mirror descent, whose gradient estimates use independent
samples. In fact, this algorithm can fail (for example, conditioning on $S'$ sharing a parity property of $S$ biases the held-out observation $Y$; see \Cref{prop:naive-amplification}).

To see what an algorithm needs instead, recall how MW is analyzed with
fresh observations. Its potential is the relative entropy $\KL(D\Vert q)$.
If an update uses a fresh observation that is independent of $q$ and of
the query $f$, it decreases the expected relative entropy from $D$ by at
least $\eta$ times the squared error of $\E_q[f]$, up to an additive
$\eta^2/8$ (\Cref{fact:mw-progress}). Thus the squared error of the answer
$\E_q[f]$ is controlled by the decrease in the potential when we perform
one more update. Over $T$ updates these decreases add up to at most the
initial relative entropy. Therefore the average squared error over the
updates is small. We turn this average into the error of our estimator
by choosing the number of updates $\mathcal T$ uniformly from
$\{0,\ldots,T-1\}$, independently of $S$.

Let $q_{t-1}(s)$ denote the distribution produced by a complete run with $t-1$ updates on a sample $s$ of size $n$. If we computed $q_{t-1}(S)$ on $S$ and then drew $(S',Y)=\Amp(S)$ for one more update, $q_{t-1}(S)$ would depend on $S$ rather than on $S'$ alone, and thus we could not apply the amplification guarantee to $(S',Y)$. We bypass this problem by reversing the order of amplification and updates: to perform $t$ updates on $S$, we \emph{first} draw $(S',Y)=\Amp(S)$, compute $q_{t-1}(S')$ from $S'$ alone (using fresh internal randomness), and apply one final MW update to $q_{t-1}(S')$ using the query $f_{S'}$ and observation $Y$. Because this $t$-update procedure accesses $S$ only through the single amplified pair $(S',Y)$, replacing $\Amp(S)$ with $n+1$ independent observations $(\bar S,\bar Y)\sim D^{n+1}$ turns it into a $(t-1)$-update run on $\bar S\sim D^n$ followed by one update with a fresh observation $\bar Y\sim D$. By the amplification guarantee, the expected potential of this $t$-update procedure is close to that of the ideal one, which the progress bound for a fresh observation controls in terms of the expected potential after $t-1$ updates. Unrolling this recursion for $\mathcal T$ steps, we first generate a chain of amplified samples from $S_{\mathcal T}=S$ down to $S_0$ via $(S_{i-1},Y_i)=\Amp(S_i)$ (for $i=\mathcal T,\ldots,1$), initialize at an estimate built from $S_0$, and then apply the MW updates in reverse order for $i=1,\ldots,\mathcal T$ using $f_{S_{i-1}}$ and $Y_i$ before answering $f_S$. Importantly, the relative entropies that telescope in the analysis are the expected potentials of complete runs of lengths $t=0,\ldots,T$ on independent samples of size $n$, rather than the steps of a single run (as in the standard analysis).

The amplification guarantee bounds the change in the expectation of a bounded function and depends linearly on the bound. Thus we need the relative entropy to remain bounded, which MW alone
does not ensure. We therefore follow each update by a KL projection onto
the distributions $q$ with $q(x)\geq1/(2d)$ for every $x$. Every such $q$
satisfies $\KL(p\Vert q)\leq\log(2d)$ for every $p\in\Delta_d$
(\Cref{fact:kl-coordinate-bound}). Moreover, the projection does not
increase the relative entropy from distributions in this convex set
(\Cref{fact:kl-projection}). Since $D$ need not belong to this set, we
estimate $\widetilde D=(D+u)/2$ instead, where $u$ is uniform on $[d]$, and
convert the answer back using $\E_D[f]=2\E_{\widetilde D}[f]-\E_u[f]$. Each
update then uses the feedback $(f(Y)+\E_u[f])/2$, whose expectation is
$\E_{\widetilde D}[f]$.
We initialize at the average of $u$ and the empirical
distribution of $S_0$. Since relative entropy on this set is at most $\log(2d)$, replacing an independent pair with an amplified pair changes the expected relative entropy after an update by at most $(\sqrt d/n)\log(2d)$ (\Cref{lem:simple-mw-recursion}). Balancing $T=\lceil n/\sqrt d\rceil$ updates with step size $\eta$ of order $d^{1/4}\sqrt{\log d/n}$ gives mean squared error $O(d^{1/4}\sqrt{\log d/n})$ (\Cref{thm:simple-mw}), as we show in \Cref{sec:efficient-estimation}.

\paragraph{A penalty around the empirical distribution.}
In this argument, each update loses $(\sqrt d/n)\log(2d)$ because $\log(2d)$
is the largest possible relative entropy. To improve this bound, we use
the second-moment guarantee for amplification
(\Cref{lem:efficient-amplification}), which bounds the change in the
expected potential by $\sqrt{2d}/n$ times the root mean square of the
potential of the ideal procedure. Our initialization is closer to
$\widetilde D$ than the worst case, since its expected relative entropy
from $\widetilde D$ is $O(\log(1+d/n))$ (\Cref{lem:mw-empirical-moments}). However, the updates can move
away from it. Thus the initialization does not suffice to bound this second
moment after every update. We therefore add to the KL projection a convex
penalty that grows with the deviation of $q$ from the average of $u$ and
the empirical distribution of the current sample. This is a composite
objective mirror descent update~\cite{COMID2010} with a penalty that
depends on the sample. We choose the penalty so that the square of the
relative entropy between two distributions in the set we project onto is at most the sum
of their penalties, and so that its expectation at $\widetilde D$ is
$O(\log^2(1+d/n))$ (\Cref{lem:mw-empirical-moments}). The penalty also
changes the progress bound, by a term that offsets the
amplification error. As a result, each update loses only
$O((\sqrt d/n)\log(1+d/n))$. With
$T=\lceil n/\sqrt d\rceil$, this gives error
$O(d^{1/4}\sqrt{\log(1+d/n)/n})$, the bound in
\Cref{thm:mirror-descent-upper-bound}, which is $O(d^{3/4}/n)$ for $n\geq d$.
We will refer to this approach as {\em penalized mirror descent with sample
amplification}. In \Cref{sec:penalized-mw} we state it for penalized MW over
any convex family of distributions (\Cref{alg:general-amplification}) and
bound its error (\Cref{thm:geometric-amplification}). We derive
\Cref{thm:mirror-descent-upper-bound} from this bound, using a penalized
projection that can be computed efficiently to the required accuracy
(\Cref{prop:penalized-mw-computation}).

\paragraph{Gaussian distributions.}
We can apply our analysis of penalized MW to any convex family of
distributions, since it relies only on an amplifier with a $\chi^2$ guarantee and a penalty
that is minimized at the initialization, bounds the square of the
potential, and has small expectation at the true distribution
(\Cref{thm:geometric-amplification}). For data from $\mathcal N(\mu,I_d)$,
we use the amplifier of~\cite[Examples~4.1 and~4.2]{axelrod2024statistical},
which keeps the empirical mean and redraws the deviations from it as
centered Gaussian noise. Its $\chi^2$ divergence from the distribution
of $n+1$ fresh observations is at most $d/n^2$ when $d\leq n^2$ and $n\geq2$
(\Cref{lem:gaussian-amplification}). If we maintained only an estimate of
the mean, the squared loss of its answer could be nonconvex. We therefore
extend the domain to mixtures of Gaussian distributions with identity
covariance, whose answers are linear in the mixing measure. We apply MW to
the density of the mixture and then project back onto mixtures with a
penalty given by the fourth moment of the mixing measure around the
empirical mean. For every selected bounded query, this gives the bound
$O(\min\{1,d^{3/4}/n\})$ in \Cref{thm:gaussian-upper-bound}, which we prove
in \Cref{sec:gaussian-estimation}. We do not give an efficient
implementation of this estimator.

To obtain an efficient estimator, we instead return to a single mean $v$
and answer with $\psi_s(v)=\E_{\mathcal N(v,I_d)}[f_s]$. Given a fresh
observation $Y\sim\mathcal N(\mu,I_d)$, the gradient step for squared loss is
\[
v\gets v-\eta\bigl(\psi_s(v)-f_s(Y)\bigr)\nabla \psi_s(v).
\]
We estimate this gradient without bias by sampling from
$\mathcal N(v,I_d)$. As with MW, we try to bound the squared answer error
by the expected decrease in a potential, now $\|v-\mu\|_2^2/2$.
Nonlinearity of $\psi_s$ makes this calculation inexact. However, Gaussian
smoothing bounds its Hessian by a universal constant, and thus a Taylor
expansion gives the desired progress
up to $O(\eta^2+\eta\|v-\mu\|_2^4)$.

To keep this error small, we project onto a ball of radius
$O(\sqrt{(d+\log n)/n})$ around the empirical mean. On independent inputs, the
ball contains $\mu$ with high probability, and projection bounds the
expected fourth power of the mean error by $O((d+\log n)^2/n^2)$.
This also bounds the second moment of the potential needed for
amplification. We can therefore use the preceding argument
with amplification and updates in reverse order, now in the Euclidean
geometry and with a constraint in place of the penalty. When $d\geq\log n$,
this gives mean squared error $O(d^{3/4}/n+d^2/n^2)$, matching the
mixture bound for $n\geq d^{5/4}$ (\Cref{thm:gaussian-sgd}).

\paragraph{Smoothed sample size.}
Finally, when the sample size is random, the sample itself provides fresh
observations, and no amplification is needed. In the smoothed variant of
the problem, the sample size is chosen uniformly from $\{n,\ldots,n+h-1\}$,
independently of the observations, and the mapping is defined on samples
of each of these sizes. For each $t$, the observation $X_{t+1}$ is
independent of the first $t$ observations. It is therefore a fresh
observation for the query selected on them and for any distribution
computed from them. We can thus perform MW updates along the prefixes of
our sample, starting from the Laplace predictor of the first $n$
observations, without amplification or projection. As with the random
number of updates, the smoothed sample size makes the error of our estimator
equal to the average error along these updates. In
\Cref{sec:random-horizon} we show that this estimator has error
$O(\sqrt{\log(1+d/n)/h})$ (\Cref{thm:random-horizon-upper-bound}).
\section{Preliminaries}
\label{sec:preliminaries}

\subsection{Notation and Probability Distances}

We write $\N=\{1,2,\ldots\}$, $[d]=\{1,\ldots,d\}$, and $\Delta_d$ for the
probability simplex on $[d]$. We write $u$ for the uniform distribution on $[d]$ and
$\widetilde\Delta_d=\{q\in\Delta_d\mid q\geq u/2\}$ for the distributions whose
coordinates are all at least $1/(2d)$. All logarithms are natural.
The symbols $\widetilde O$ and $\widetilde\Omega$ suppress logarithmic
factors in $n$ and $d$.

For $D\in\Delta_d$, $A\subseteq[d]$, and $f\colon[d]\to\mathbb R$, we write
$D(A)=\sum_{x\in A}D(x)$, $\E_D[f]=\sum_xD(x)f(x)$, and
$\supp(D)=\{x\mid D(x)>0\}$.
When using vector notation, we identify $f$ with $(f(1),\ldots,f(d))$
and $D$ with $(D(1),\ldots,D(d))$.
We use $\1\{E\}$ for the indicator of an event, $\1_A$ for the indicator vector of a set $A$, and $\1$ for the vector of all ones. The empirical distribution
of $S=(X_1,\ldots,X_n)$ is
$\widehat D_S(x)=n^{-1}\sum_{i=1}^n\1\{X_i=x\}$.
The norms $\|\cdot\|_1$, $\|\cdot\|_2$, and $\|\cdot\|_\infty$ are
the usual vector norms. We omit the subscript for the Euclidean norm
($\|v\|=\|v\|_2$) and use $\langle v,w\rangle=v^\top w$ for the
Euclidean inner product. Superscript $\top$ denotes transpose.

We denote the product of two distributions
by $\mu\otimes\nu$, and for a randomized procedure $\Amp$ and distribution $D$, we write $\Amp(D^n)$ for the distribution of $\Amp(S)$ when $S\sim D^n$. For distributions $\mu,\nu$ with densities $f,g$,
total variation, relative entropy, and $\chi^2$ divergence are
\begin{align*}
\TV(\mu,\nu)&=\frac12\int|f-g|,
&\KL(\mu\Vert\nu)&=\int f\log\frac fg,
&\chi^2(\mu\Vert\nu)&=\int\left(\frac fg-1\right)^2g.
\end{align*}
For discrete distributions the integrals are sums. A positive numerator
with a zero denominator gives infinite divergence.

\begin{fact}[Total variation and expectations]
\label[fact]{fact:tv-expectation}
For any probability distributions $\mu,\nu$,
\[
\TV(\mu,\nu)=\sup_{0\leq f\leq1}|\E_\mu[f]-\E_\nu[f]|.
\]
Consequently, if $0\leq f\leq B$, then
$|\E_\mu[f]-\E_\nu[f]|\leq B\TV(\mu,\nu)$.
Applying the same randomized procedure to both distributions cannot
increase their total variation distance.
\end{fact}

\begin{fact}[Change of measure by Cauchy--Schwarz]
\label[fact]{fact:chi-square-change-of-measure}
If $\chi^2(\mu\Vert\nu)<\infty$ and $\E_\nu[f^2]<\infty$, then
\[
|\E_\mu[f]-\E_\nu[f]|
\leq\sqrt{\chi^2(\mu\Vert\nu)\E_\nu[f^2]}.
\]
If $0\leq f\leq B$, the right-hand side is at most
$\sqrt{\chi^2(\mu\Vert\nu)B\E_\nu[f]}$.
The comparison also holds after applying the same randomized procedure
to both inputs.
\end{fact}

\begin{fact}[Bounded relative entropy]
\label[fact]{fact:kl-coordinate-bound}
If $q\in\Delta_d$ satisfies $q(x)\geq a>0$ for every $x$, then
$0\leq\KL(p\Vert q)\leq\log(1/a)$ for every $p\in\Delta_d$.
\end{fact}

\subsection{Sample Amplification}

For $s\in[d]^n$, we append a copy of a uniformly chosen entry and
uniformly shuffle the resulting $n+1$ samples, writing
$\Amp(s)=(s',y)$ for the first $n$ entries and the last entry.
This takes $O(n)$ operations. Our construction follows the resampling
approach to discrete sample amplification
\cite{axelrod2020amplification,axelrod2024statistical}. We calculate its
$\chi^2$ divergence when adding one sample in \Cref{app:amplification};
the expectation bounds below then follow from
\Cref{fact:tv-expectation,fact:chi-square-change-of-measure}.

\begin{lemma}[Amplification by one sample]
\label[lemma]{lem:efficient-amplification}
For every $n\geq1$, $d\geq2$, and $D\in\Delta_d$,
\[
\TV\bigl(\Amp(D^n),D^{n+1}\bigr)\leq\frac{\sqrt d}{n},
\qquad
\chi^2\bigl(\Amp(D^n)\Vert D^{n+1}\bigr)
\leq\frac{2d}{n^2}.
\]
Consequently, for $S\sim D^n$, $(\bar S,\bar Y)\sim D^{n+1}$, and any randomized procedure $F$ whose internal coins are independent of its input and have the same distribution in both experiments, we have
\[
\begin{aligned}
\left|\E F(\Amp(S))-\E F(\bar S,\bar Y)\right|
&\leq \frac{L\sqrt d}{n} &&\text{if }0\leq F\leq L,\\
\left|\E F(\Amp(S))-\E F(\bar S,\bar Y)\right|
&\leq\frac{\sqrt{2d}}{n}\sqrt{\E F(\bar S,\bar Y)^2},
\end{aligned}
\]
where the second bound assumes $\E F(\bar S,\bar Y)^2<\infty$.
\end{lemma}

\subsection{Multiplicative Weights and KL Projection}

For a positive distribution $r$, $f\colon[d]\to[0,1]$, $b\in[0,1]$,
and $\eta>0$, we write
\[
\MW_{\eta}(r,f,b)(x)\propto
r(x)\exp\{\eta(b-\E_r[f])f(x)\}.
\]
This specializes the exponential update in
\cite[Sections~2.1--2.2]{arora2012multiplicative} to squared loss.
Applying Hoeffding's lemma~\cite{Hoeffding1963} ($\log\E_{X\sim r}[e^{t(f(X)-\E_r[f])}]\leq t^2/8$) to its normalizing
constant proves the following progress bound.

\begin{fact}[MW progress]
\label[fact]{fact:mw-progress}
For $r,f,b,\eta$ as above and every $w\in\Delta_d$,
\[
\KL\bigl(w\Vert\MW_{\eta}(r,f,b)\bigr)
\leq\KL(w\Vert r)
+\eta(b-\E_r[f])(\E_r[f]-\E_w[f])
+\frac{\eta^2(b-\E_r[f])^2}{8}.
\]
In particular, if $b=f(Y)$ for an independent $Y\sim D$, then
\[
\E_Y\KL\bigl(D\Vert\MW_{\eta}(r,f,f(Y))\bigr)
\leq\KL(D\Vert r)-\eta(\E_r[f]-\E_D[f])^2+\eta^2/8.
\]
The first inequality also holds for probability densities on a common
measure space, with integrals in place of sums, whenever $\KL(w\Vert r)$
is finite and $f$ takes values in $[0,1]$.
\end{fact}

KL projection cannot increase relative entropy from a distribution in the
feasible set. We use the following standard inequality for the simple update.

\begin{fact}[KL projection {\cite[Section~2.2]{arora2012multiplicative}}]
\label[fact]{fact:kl-projection}
Let $\mathcal C\subseteq\Delta_d$ be a nonempty closed convex set of
positive distributions, and let $z\in\Delta_d$ have positive coordinates.
If $v=\argmin_{r\in\mathcal C}\KL(r\Vert z)$, then
\[
\KL(w\Vert v)+\KL(v\Vert z)\leq\KL(w\Vert z)
\qquad\text{for every }w\in\mathcal C.
\]
\end{fact}

\Needspace{18\baselineskip}
For the update in \Cref{sec:penalized-mw}, we add a convex penalty and
allow optimization error. We use the optimality condition for composite
objective mirror descent \cite[Equation~(5)]{COMID2010} to account for
both changes.

\begin{corollary}[Penalized KL projection]
\label[corollary]{cor:penalized-kl-projection}
Let $\mathcal C,z$ be as in \Cref{fact:kl-projection}, and let $P$ be
a continuously differentiable convex function on an open set containing
$\mathcal C$. For $\lambda\geq0$, define
\[
v=\argmin_{r\in\mathcal C}
\bigl\{\KL(r\Vert z)+\lambda P(r)\bigr\}.
\]
For every $w\in\mathcal C$,
\[
\KL(w\Vert v)+\KL(v\Vert z)+\lambda P(v)
\leq\KL(w\Vert z)+\lambda P(w).
\]
More generally, for $\tau\geq0$, if $v\in\mathcal C$ satisfies
\[
\max_{w\in\mathcal C}
\left\langle\nabla_v\bigl(\KL(v\Vert z)+\lambda P(v)\bigr),v-w\right\rangle
\leq\tau,
\]
the inequality holds with $\tau$ added to the right-hand side.
\end{corollary}

\begin{proof}
For every $w\in\mathcal C$, convexity of $P$ implies
\[
\begin{aligned}
&\KL(w\Vert v)+\KL(v\Vert z)-\KL(w\Vert z)
+\lambda(P(v)-P(w))\\
&\qquad\leq
\left\langle\log\frac{v}{z}+\lambda\nabla P(v),v-w\right\rangle
\leq\tau.
\end{aligned}
\]
For an exact minimizer, the optimality condition holds with $\tau=0$.
\end{proof}

\begin{fact}[Computing KL projection {\cite[Theorem~7]{HerbsterWarmuth2001}}]
\label[fact]{fact:kl-projection-computation}
Let $z\in\Delta_d$ have positive coordinates.
The minimizer of $\KL(q\Vert z)$ over $q\in\widetilde\Delta_d$ has
coordinates
\[
q(x)=\max\left\{\frac1{2d},cz(x)\right\},
\qquad c>0,\quad \sum_xq(x)=1.
\]
This projection can be computed in $O(d)$ operations.
\end{fact}
\iffulldraft
\section{A Simple Estimator Using Multiplicative Weights}
\label{sec:efficient-estimation}

We prove the simple MW bound described in \Cref{sec:overview} by generating a
chain of amplified samples and applying projected multiplicative weights
updates in reverse order. The projection bounds relative entropy, and thus the
total variation guarantee for amplification lets us transfer the progress
of an update with a fresh observation to this procedure. In
\Cref{sec:penalized-mw}, we improve the bound by adding a penalty to the
projection.

We count one evaluation of $\dde$ as obtaining the vector
$(\dde(s)(x))_{x\in[d]}$ for a sample $s\in[d]^n$. If access is through
individual values $\dde(s)(x)$, obtaining this vector costs $d$
evaluations. This convention also applies to \Cref{sec:penalized-mw,sec:random-horizon}.

\begin{theorem}[Simple MW estimator]
\label[theorem]{thm:simple-mw}
For every $n\geq1$, $d\geq2$, and mapping $\dde$ from
$[d]^n$ to $[0,1]^{[d]}$, there is a randomized estimator $\phat$ satisfying
\[
\sup_{D\in\Delta_d}\E_{S\sim D^n}
\bigl[(\phat(S)-\E_D[\dde(S)])^2\bigr]
=O\left(\min\left\{1,\frac{d^{1/4}}{\sqrt n}\sqrt{\log d}\right\}\right).
\]
It uses at most $T=\lceil n/\sqrt d\rceil$ evaluations of $\dde$ and
$O(T(n+d))$ additional operations.
\end{theorem}

As in \Cref{sec:overview}, let $f_s=\dde(s)$. We will bound mean
squared error by the decrease in expected relative entropy between
complete runs with $t$ and $t+1$ updates on independent inputs,
where $q_t(s)$ denotes the output distribution of a run with $t$ updates
on input $s$. Choosing the number of updates uniformly will then turn
the resulting average error bound into the risk of our estimator.
To apply the total variation guarantee for amplification, we keep relative
entropy bounded by restricting our estimates to $\widetilde\Delta_d$. For any $q$ in this set, $\KL(D\Vert q)\leq\log(2d)$.
The KL projection inequality also requires the distribution we are
estimating to belong to $\widetilde\Delta_d$. We therefore estimate
$\widetilde D$ and initialize at $e_s$, where
\[
\widetilde D=\frac{D+u}{2},\qquad
e_s=\frac{\widehat D_s+u}{2}.
\]
An estimate of $\E_{\widetilde D}[f]$ then lets us estimate $\E_D[f]$.
To see this, for every $q\in\widetilde\Delta_d$ and $f\in[0,1]^{[d]}$,
$2\E_q[f]-\E_u[f]\in[0,1]$ and
\begin{equation}
\bigl(2\E_q[f]-\E_u[f]-\E_D[f]\bigr)^2
=4\bigl(\E_q[f]-\E_{\widetilde D}[f]\bigr)^2.
\label{eq:mw-mixed-answer}
\end{equation}

For fixed $q$ and $f$ and a step size $\eta>0$, an MW update with feedback of expectation
$\E_{\widetilde D}[f]$ decreases expected relative entropy from
$\widetilde D$ by $\eta$ times the squared error of its answer, up to
the quadratic remainder in \Cref{fact:mw-progress}.
For $q\in\widetilde\Delta_d$, $f\in[0,1]^{[d]}$, and $y\in[d]$,
we set $b=(f(y)+\E_u[f])/2$ and define
\begin{equation}
q^+=\MW_\eta(q,f,b),\qquad
\mathsf U(q,f,b)=\argmin_{v\in\widetilde\Delta_d}\KL(v\Vert q^+).
\label{eq:simple-mw-projection}
\end{equation}

In \Cref{alg:simple-mw}, we generate the amplified samples from
$S_{\mathcal T}=S$ down to $S_0$, then apply
\eqref{eq:simple-mw-projection} for $i=1,\ldots,\mathcal T$.
Thus the updates run in the reverse order of amplification. This order
ensures that the distribution to which each update is applied is itself
the output of a complete run on the sample that selects its query.
Each amplification uses fresh randomness, independently of $S$ and the
chosen number of updates. In \Cref{app:naive-amplification} we show that amplifying the original sample before every update instead can fail (\Cref{prop:naive-amplification}).

\begin{algorithm}[H]
\caption{Simple MW estimator}
\label{alg:simple-mw}
\small
\begin{algorithmic}[1]
\Statex \textbf{Input} $S\in[d]^n$, mapping $\dde$, step size $\eta>0$, integer $T\geq1$
\State Choose $\mathcal T$ uniformly from $\{0,\ldots,T-1\}$, independently of $S$
\State $S_{\mathcal T}\gets S$
\For{$i=\mathcal T,\mathcal T-1,\ldots,1$}
    \State $(S_{i-1},Y_i)\gets\Amp(S_i)$
    \State $f_i\gets\dde(S_{i-1})$, \quad $b_i\gets\bigl(f_i(Y_i)+\E_u[f_i]\bigr)/2$
\EndFor
\Statex
\State $q\gets e_{S_0}$
\For{$i=1,\ldots,\mathcal T$}
    \State $q\gets\mathsf U(q,f_i,b_i)$
\EndFor
\State $f\gets\dde(S)$
\State \Return $2\E_q[f]-\E_u[f]$
\end{algorithmic}
\end{algorithm}

More precisely, $q_t(s)$ is the final distribution $q$ in a complete run of
\Cref{alg:simple-mw} with $t$ MW updates on input $s$. In a run with
$\mathcal T$ updates, the state after $i$ updates is $q_i(S_i)$, using
the randomness from the generated chain. If $\mathcal T=0$, both loops
are empty and $S_0=S$. In particular, $q_0(s)=e_s$, and for $t\geq0$,
\begin{equation}
\begin{aligned}
(s',y)&=\Amp(s),\qquad f'=\dde(s'),\qquad
b'=\tfrac12(f'(y)+\E_u[f']),\\
q_{t+1}(s)&=\mathsf U\bigl(q_t(s'),f',b'\bigr).
\end{aligned}
\label{eq:mw-depth-recursion}
\end{equation}
To compute $q_t(s')$, we use only $s'$ and fresh internal randomness,
performing its $t$ updates before the final update using $f'$ and $y$.

The projection in \eqref{eq:simple-mw-projection} preserves the progress
of an update with a fresh observation, because
$\widetilde D\in\widetilde\Delta_d$.

\begin{corollary}[Projected MW with a fresh sample]
\label[corollary]{cor:projected-mw-progress}
Let $\eta>0$, $D\in\Delta_d$, $q\in\widetilde\Delta_d$, and $f\in[0,1]^{[d]}$.
For $Y\sim D$, let $b(Y)=(f(Y)+\E_u[f])/2$. Then
\begin{equation}
\begin{aligned}
\E_Y\KL\!\left(\widetilde D\,\middle\Vert\,
\mathsf U(q,f,b(Y))\right)
&\leq\KL(\widetilde D\Vert q)\\
&\quad-\eta(\E_q[f]-\E_{\widetilde D}[f])^2+\eta^2/32.
\end{aligned}
\label{eq:projected-mw-fresh}
\end{equation}
\end{corollary}

\begin{proof}
Let $b=b(Y)$. Both $b$ and $\E_q[f]$ belong to
$[\E_u[f]/2,(1+\E_u[f])/2]$. By
\Cref{fact:mw-progress,fact:kl-projection}, for every
$w\in\widetilde\Delta_d$,
\begin{equation}
\KL(w\Vert\mathsf U(q,f,b))
\leq\KL(w\Vert q)
+\eta(b-\E_q[f])(\E_q[f]-\E_w[f])+\eta^2/32.
\label{eq:simple-mw-pointwise}
\end{equation}
Taking $w=\widetilde D$ and averaging over $Y$ (using $\E_Y[b(Y)]=\E_{\widetilde D}[f]$), we obtain \eqref{eq:projected-mw-fresh}.
\end{proof}

The samples along the chain generated by \Cref{alg:simple-mw} are
dependent. Our main technical lemma
therefore relates the expected potentials of complete runs with
different numbers of updates, each evaluated on an independent input, rather
than telescoping the potentials along one chain.

\begin{lemma}
\label[lemma]{lem:simple-mw-recursion}
For every $\eta>0$ and $D\in\Delta_d$, define
\begin{equation}
K_t=\E_{S\sim D^n}\KL(\widetilde D\Vert q_t(S)),\qquad
\ell_t=\E_{S\sim D^n}
\bigl(\E_{q_t(S)}[f_S]-\E_{\widetilde D}[f_S]\bigr)^2,
\label{eq:mw-depth-potentials}
\end{equation}
where expectations are over $S\sim D^n$ and the randomness of the
complete run defining $q_t(S)$. Then, for every $t\geq0$,
\begin{equation}
\eta\ell_t\leq K_t-K_{t+1}+\eta^2/32+\frac{\sqrt d}{n}\log(2d).
\label{eq:simple-mw-recursion}
\end{equation}
We can compute $q_t(s)$ using $t$ evaluations of $\dde$ and
$O((t+1)(n+d))$ additional operations.
\end{lemma}

We first deduce \Cref{thm:simple-mw} from this lemma.

\begin{proof}[Proof of \Cref{thm:simple-mw}]
\Cref{alg:simple-mw} returns
\[
\phat(S)=2\E_{q_{\mathcal T}(S)}[f_S]-\E_u[f_S].
\]
By the answer conversion \eqref{eq:mw-mixed-answer}, its risk is
$4T^{-1}\sum_{t=0}^{T-1}\ell_t$. Summing
\eqref{eq:simple-mw-recursion} and using
$K_0\leq\log(2d)$ and $K_T\geq0$, we obtain
\begin{equation}
\E\bigl[(\phat(S)-\E_D[\dde(S)])^2\bigr]
\leq4\left(\frac{\log(2d)/T+(\sqrt d/n)\log(2d)}{\eta}
+\frac\eta{32}\right).
\label{eq:simple-mw-bound}
\end{equation}
Taking $T=\lceil n/\sqrt d\rceil$ and
$\eta=\sqrt{(\sqrt d/n)\log(2d)}$ bounds this expression by
$O(d^{1/4}\sqrt{\log d/n})$. Since $\phat(S)\in[0,1]$, its risk is
also at most one. Computing $q_{\mathcal T}(S)$ and evaluating $\dde(S)$
uses at most $T$ mapping evaluations and $O(T(n+d))$ additional operations.
\end{proof}

It remains to prove \eqref{eq:simple-mw-recursion}. We first run $t$
updates on an independent sample and apply one more update with an independent
observation. \Cref{cor:projected-mw-progress} bounds the expected
potential of this procedure by $K_t-\eta\ell_t+\eta^2/32$. Replacing
the independent sample and fresh observation with one amplified pair makes the
procedure a complete run with $t+1$ updates, whose expected potential is
$K_{t+1}$.
Since projection keeps relative entropy bounded by $\log(2d)$, replacing
independent inputs with amplified inputs changes its expectation by at most
$(\sqrt d/n)\log(2d)$.

\begin{proof}[Proof of \Cref{lem:simple-mw-recursion}]
Every update returns a distribution in $\widetilde\Delta_d$.

\paragraph{Progress with an independent sample.}
For fixed $D$ and $t$, we draw $(\bar S,\bar Y)\sim D^{n+1}$ and let
\[
\bar f=\dde(\bar S),\qquad
V=\mathsf U\!\left(q_t(\bar S),\bar f,
\frac{\bar f(\bar Y)+\E_u[\bar f]}2\right).
\]
Conditional on $\bar S$ and $q_t(\bar S)$, we apply
\Cref{cor:projected-mw-progress} with $q=q_t(\bar S)$ and $f=\bar f$.
Averaging over $\bar S$ and the randomness of $q_t$, we obtain
\begin{equation}
\E\KL(\widetilde D\Vert V)
\leq K_t-\eta\ell_t+\eta^2/32.
\label{eq:simple-mw-independent}
\end{equation}

\paragraph{Amplification.}
We apply \Cref{lem:efficient-amplification} to the randomized function
\[
G(s,y)=\KL\!\left(\widetilde D\,\middle\Vert\,
\mathsf U\bigl(q_t(s),f_s,
\tfrac12(f_s(y)+\E_u[f_s])\bigr)\right).
\]
On independent inputs $(\bar S,\bar Y)$, its expectation is
$\E\KL(\widetilde D\Vert V)$. By the update recursion \eqref{eq:mw-depth-recursion}, its
expectation on $\Amp(S)$, for $S\sim D^n$, is $K_{t+1}$.
Moreover, $0\leq G\leq\log(2d)$. Thus
\begin{equation}
K_{t+1}\leq\E\KL(\widetilde D\Vert V)+\frac{\sqrt d}{n}\log(2d).
\label{eq:simple-mw-amplified}
\end{equation}
Combining \eqref{eq:simple-mw-independent} and
\eqref{eq:simple-mw-amplified} proves \eqref{eq:simple-mw-recursion}.

\paragraph{Running time.}
With the number of updates fixed to $t$, \Cref{alg:simple-mw} implements
\eqref{eq:mw-depth-recursion} using $t$ mapping evaluations.
Amplification takes $O(n)$ operations per call, while the multiplicative
weights update and KL projection together take $O(d)$ by
\Cref{fact:kl-projection-computation}. Including initialization, this is
$O((t+1)(n+d))$ operations in total.
\end{proof}
\section{Penalized Mirror Descent with Sample Amplification}
\label{sec:penalized-mw}

We now prove \Cref{thm:mirror-descent-upper-bound} by replacing the worst-case
bound $\log(2d)$ on relative entropy with a second-moment bound. For $n\geq d$,
this improves the bound of \Cref{thm:simple-mw} from
$O(d^{1/4}\sqrt{\log d/n})$ to $O(d^{3/4}/n)$. As described in
\Cref{sec:overview}, the initialization $e_s$ gives a good localization of
$\widetilde D$, but the updates can move away from it.
The standard solution is to introduce a convex constraint on the iterates that
restricts them to stay near $e_s$ and enforce the constraint
via a KL projection, as in Section~\ref{sec:efficient-estimation}. However, due
to the randomness in $e_s$, the constraint based on it may
be violated by the true distribution. The extent of violation can be controlled
using McDiarmid's concentration inequality. This simple argument can provide the guarantees of \Cref{thm:mirror-descent-upper-bound}
with extra logarithmic factors to account for the constraint violations.

In this section, we provide a tighter analysis which bypasses the constraint violation
analysis by transferring the hard convex constraint to a softer convex penalty around $e_s$
added to the KL projection in \Cref{alg:simple-mw}. This leads to an analysis in which only the expected
convex penalty matters, allowing us to avoid the superfluous logarithmic factors. We capture this approach through a general framework based on penalized MW over any convex family of distributions
(\Cref{sec:mirror-descent-amplification}) which we present below. We also use the general framework for Gaussian mixtures in \Cref{sec:gaussian-mixtures}. In \Cref{sec:penalized-mw-proof} we construct the penalty for discrete distributions and prove \Cref{thm:mirror-descent-upper-bound}.

\subsection{The General Framework}
\label{sec:mirror-descent-amplification}

Let $\mathcal K$ be a convex set of probability densities on a sample space $\Xcal$ with respect to a common reference measure (for discrete distributions in \Cref{sec:efficient-estimation,sec:penalized-mw-proof}, $\Xcal=[d]$ and $\mathcal K=\widetilde\Delta_d$).
An amplifier maps a sample $s\in\Xcal^n$ to a pair
$\Amp(s)=(s',y)\in\Xcal^n\times\Xcal$, using independent randomness (for
discrete distributions, the amplifier of \Cref{lem:efficient-amplification}).
Each sample $s\in\Xcal^n$ determines an initialization $q_0(s)\in\mathcal K$ (in \Cref{alg:simple-mw}, $e_s$), a query $f_s\colon\Xcal\to[0,1]$, feedback $b_s(y)$ from an observation $y$ (in \Cref{alg:simple-mw}, $(f_s(y)+\E_u[f_s])/2$), and a penalty $P_s\colon\mathcal K\to[0,\infty)$.
For a density $r$, we write $\MW_\eta(r,f,b)$ for the density proportional to
$r\exp\{\eta(b-\E_r[f])f\}$, as in \Cref{sec:preliminaries}. With step size
$\eta>0$ and penalty weight
$\lambda\geq0$, the exact update of $r$ is the penalized KL projection of the MW step,
\begin{equation}
\mathsf U_s(r,y)=\argmin_{v\in\mathcal K}\left\{\KL\bigl(v\,\Vert\,
\MW_\eta(r,f_s,b_s(y))\bigr)+\lambda P_s(v)\right\},
\label{eq:penalized-mw-update}
\end{equation}
and we also allow $\mathsf U_s(r,y)\in\mathcal K$ to be an approximate minimizer satisfying condition~(H4) below.
Without a penalty ($\lambda=0$), this is the projected MW update
\eqref{eq:simple-mw-projection} of \Cref{sec:efficient-estimation}. For
\Cref{thm:mirror-descent-upper-bound}, we add a penalty centered at $e_s$, which
we construct in \Cref{sec:penalized-mw-proof}. For a fixed sample, the
penalized update is a composite objective mirror descent step
\cite[Eq.~(3)]{COMID2010}. In \Cref{sec:gaussian-sgd}, we use stochastic
gradient steps instead, which are mirror descent steps in the Euclidean
geometry. All procedures are fixed independently of the unknown data distribution $D$, and when $\Xcal$ is continuous they are measurable.

\begin{algorithm}[H]
\caption{Penalized MW with sample amplification}
\label{alg:general-amplification}
\small
\begin{algorithmic}[1]
\Statex \textbf{Input} $S\in\Xcal^n$, step size $\eta>0$, penalty weight $\lambda\geq0$, integer $T\geq1$
\Statex \textbf{Given} amplifier $\Amp$, initialization $q_0$, queries $f_s$, feedback $b_s$, penalties $P_s$
\State Choose $\mathcal T$ uniformly from $\{0,\ldots,T-1\}$, independently of $S$
\State $S_{\mathcal T}\gets S$
\For{$i=\mathcal T,\mathcal T-1,\ldots,1$}
    \State $(S_{i-1},Y_i)\gets\Amp(S_i)$
\EndFor
\Statex
\State $r\gets q_0(S_0)$
\For{$i=1,\ldots,\mathcal T$}
    \State $r\gets\mathsf U_{S_{i-1}}(r,Y_i)$
\EndFor
\State \Return $\E_r[f_S]$
\end{algorithmic}
\end{algorithm}

With the discrete choices above and $\lambda=0$,
\Cref{alg:general-amplification} is \Cref{alg:simple-mw}, except that
\Cref{alg:simple-mw} converts its answer via the answer conversion \eqref{eq:mw-mixed-answer}. As in
\Cref{sec:efficient-estimation}, let $q_t(s)$
be the output of a complete run with $t$ updates on input $s$. Then $q_0(s)$ is
the initialization and, as in the update recursion \eqref{eq:mw-depth-recursion},
\begin{equation}
q_{t+1}(s)=\mathsf U_{s'}\bigl(q_t(s'),y\bigr),\qquad(s',y)=\Amp(s),
\label{eq:general-depth-recursion}
\end{equation}
where $q_t(s')$ is computed from $s'$ and fresh internal randomness.

The proof of \Cref{lem:simple-mw-recursion} compares the last update of a run
with the same update applied to an independent sample and observation. The
following lemma makes this comparison using the $\chi^2$ guarantee for
amplification. It uses only that the update depends on the amplified pair and
on independent randomness, and we also use it in \Cref{sec:gaussian-sgd}.

\begin{lemma}[Transfer to amplified samples]
\label[lemma]{lem:amplification-transfer}
Let $D$ be a distribution on $\Xcal$ and let $\gamma>0$ satisfy
\[
\chi^2\bigl(\Amp(D^n)\Vert D^{n+1}\bigr)\leq\gamma^2.
\]
Suppose that the outputs $q_t(s)$ of complete runs satisfy
\eqref{eq:general-depth-recursion} for an update $\mathsf U_s(r,y)$ whose
internal randomness is independent of its inputs. For $t\geq0$, $S\sim D^n$, and
$(\bar S,\bar Y)\sim D^{n+1}$, let
$V_t=\mathsf U_{\bar S}(q_t(\bar S),\bar Y)$. Then every nonnegative function
$G$ of the output satisfies
\[
\E G(q_{t+1}(S))\leq\E G(V_t)+\gamma\sqrt{\E G(V_t)^2}.
\]
\end{lemma}

\begin{proof}
The quantity $G(\mathsf U_{s'}(q_t(s'),y))$ is a randomized function of the
pair $(s',y)$, and its internal randomness is independent of the pair. By
\eqref{eq:general-depth-recursion}, its expectation is $\E G(q_{t+1}(S))$ on
$\Amp(S)$ and $\E G(V_t)$ on $(\bar S,\bar Y)$. The claim follows from
\Cref{fact:chi-square-change-of-measure}.
\end{proof}

We now bound the error of \Cref{alg:general-amplification} for any data distribution $D$ on $\Xcal$ and comparator density $\theta\in\mathcal K$ under four
conditions, which we verify for discrete distributions in
\Cref{sec:penalized-mw-proof} and for Gaussian mixtures in
\Cref{sec:gaussian-mixtures}. In each ideal step on $(\bar S,\bar Y)\sim D^{n+1}$, the penalized projection subtracts $\lambda\E\KL(\theta\Vert V_t)^2$ from the expected potential while adding at most $2\lambda\E P_{\bar S}(\theta)\leq2\lambda R^2$, by the bounds $\KL(\theta\Vert V_t)^2\leq P_{\bar S}(\theta)+P_{\bar S}(V_t)$ and $\E P_S(\theta)\leq R^2$ in (H3) below. Transferring to an amplified pair via \Cref{lem:amplification-transfer} adds at most $\gamma\sqrt{\E\KL(\theta\Vert V_t)^2}$, which the subtracted term absorbs up to $\gamma^2/(4\lambda)$. Thus the analysis needs no bound on $\E\KL(\theta\Vert V_t)^2$.

\begin{theorem}[Penalized MW with sample amplification]
\label[theorem]{thm:geometric-amplification}
Let $D$ be a distribution on $\Xcal$, let $\theta\in\mathcal K$, let $\gamma,R,\lambda>0$ and $\tau\geq0$, and suppose that the following hold for every sample $s\in\Xcal^n$, observation $y\in\Xcal$, and $r\in\mathcal K$.
\begin{enumerate}[label=(H\arabic*)]
\item \emph{Amplification}:
$\chi^2(\Amp(D^n)\Vert D^{n+1})\leq\gamma^2$.
\item \emph{Feedback}: $b_s(y)\in[0,1]$ and $\E_{Y\sim D}[b_s(Y)]=\E_\theta[f_s]$.
\item \emph{Penalty}: the initialization satisfies $q_0(s)\in\mathcal K$ and
$P_s(q_0(s))=0$, all $p,q\in\mathcal K$ satisfy
$\KL(p\Vert q)^2\leq P_s(p)+P_s(q)$, and
$\E_{S\sim D^n}P_S(\theta)\leq R^2$.
\item \emph{Approximate projection}: for $z=\MW_\eta(r,f_s,b_s(y))$, the update
$v=\mathsf U_s(r,y)\in\mathcal K$ satisfies
$\KL(w\Vert v)+\lambda P_s(v)\leq\KL(w\Vert z)+\lambda P_s(w)+\tau$ for every
$w\in\mathcal K$.
\end{enumerate}
For $t\geq0$, define
\[
K_t=\E_{S\sim D^n}\KL(\theta\Vert q_t(S)),\qquad
\ell_t=\E_{S\sim D^n}\bigl(\E_{q_t(S)}[f_S]-\E_\theta[f_S]\bigr)^2,
\]
where the expectations include the randomness of the complete run defining
$q_t(S)$. Then
\begin{equation}
\eta\ell_t\leq K_t-K_{t+1}+\frac{\eta^2}8+2\lambda R^2
+\frac{\gamma^2}{4\lambda}+\tau.
\label{eq:geometric-recursion}
\end{equation}
In particular, for every $T\geq1$, the output of
\Cref{alg:general-amplification} satisfies
\begin{equation}
\E_{S\sim D^n}
\bigl[\bigl(\E_{q_{\mathcal T}(S)}[f_S]-\E_\theta[f_S]\bigr)^2\bigr]
\leq\frac R{\eta T}+\frac\eta8
+\frac{2\lambda R^2+\gamma^2/(4\lambda)+\tau}\eta,
\label{eq:geometric-master}
\end{equation}
and for $\lambda=\gamma/(2R)$, $T\geq1/(2\gamma)$,
$\eta=\sqrt{\gamma R}$, and $\tau\leq\gamma R$, this error is $O(\sqrt{\gamma R})$.
\end{theorem}

\begin{proof}
Since $P_S(q_0(S))=0$, (H3) and Jensen's inequality
give $K_0\leq\sqrt{\E P_S(\theta)}\leq R$. For $t\geq0$ with $K_t<\infty$, we draw $(\bar S,\bar Y)\sim D^{n+1}$ and let
\[
f=f_{\bar S},\qquad r=q_t(\bar S),\qquad b=b_{\bar S}(\bar Y),\qquad
V_t=\mathsf U_{\bar S}(r,\bar Y),\qquad W_t=\E\KL(\theta\Vert V_t)^2.
\]
By (H4) with $w=\theta$, and by \Cref{fact:mw-progress} with
$|b-\E_r[f]|\leq1$,
\[
\KL(\theta\Vert V_t)+\lambda P_{\bar S}(V_t)
\leq\KL(\theta\Vert r)+\eta(b-\E_r[f])(\E_r[f]-\E_\theta[f])
+\frac{\eta^2}8+\lambda P_{\bar S}(\theta)+\tau.
\]
Conditional on $\bar S$ and the randomness used to compute $r$, the
observation $\bar Y$ has distribution $D$, and thus $\E[b\mid\bar S,r]=\E_\theta[f]$ by
(H2). By (H3), we also have $\lambda P_{\bar S}(V_t)\geq\lambda\KL(\theta\Vert V_t)^2-\lambda P_{\bar S}(\theta)$. Substituting this lower bound, taking expectations, and using $\E P_{\bar S}(\theta)\leq R^2$, we obtain
\begin{equation}
\E\KL(\theta\Vert V_t)+\lambda W_t
\leq K_t-\eta\ell_t+\frac{\eta^2}8+2\lambda R^2+\tau.
\label{eq:geometric-independent-progress}
\end{equation}

Thus $W_t<\infty$. By (H1) and \Cref{lem:amplification-transfer} with $G=\KL(\theta\Vert\cdot)$, we have $K_{t+1}\leq\E\KL(\theta\Vert V_t)+\gamma\sqrt{W_t}$. Since $\gamma\sqrt{W_t}-\lambda W_t\leq\gamma^2/(4\lambda)$, combining this bound with \eqref{eq:geometric-independent-progress} proves \eqref{eq:geometric-recursion}, and in particular $K_{t+1}<\infty$.

Summing \eqref{eq:geometric-recursion} over $t=0,\ldots,T-1$ and using $K_T\geq0$
and $K_0\leq R$, we get \eqref{eq:geometric-master} because $\mathcal T$ is uniform on $\{0,\ldots,T-1\}$ independently of $S$ and the left-hand side of \eqref{eq:geometric-master} equals $T^{-1}\sum_{t<T}\ell_t$. For the
stated parameters, $2\lambda R^2+\gamma^2/(4\lambda)+\tau\leq\tfrac52\gamma R$ and
$R/T\leq2\gamma R$; thus the right-hand side of \eqref{eq:geometric-master} is
at most $\tfrac92\gamma R/\eta+\eta/8=O(\sqrt{\gamma R})$.
\end{proof}

\subsection{Improved Bound for Discrete Distributions}
\label{sec:penalized-mw-proof}

We now apply the framework to discrete distributions and prove our main
result, which we restate for convenience.

\mirrordescentupperbound*

Our estimator adds a penalty around the empirical distribution to each KL
projection in \Cref{alg:simple-mw}. Formally, it is
\Cref{alg:general-amplification} with the family $\widetilde\Delta_d$, the
initialization $e_s$, and the feedback $(f_s(y)+\E_u[f_s])/2$ of
\Cref{alg:simple-mw}, together with the penalty $P_s$ defined below, where each update \eqref{eq:penalized-mw-update} is computed up to approximation error $\tau$ in the first-order optimality condition \eqref{eq:penalized-mw-approx-opt} via \Cref{prop:penalized-mw-computation}. As in
\Cref{alg:simple-mw}, it returns $2\E_r[f_S]-\E_u[f_S]$ for its final
distribution $r$. The penalty in update~$i$ is centered at $e_{S_{i-1}}$, for the sample $S_{i-1}$ that selects the query of
this update. To compute the projection, we also keep $e_{S_{i-1}}$ from each
amplification step.

We now construct a penalty that satisfies (H3) with $R=O(\log(1+d/n))$. For
$p,q\in\widetilde\Delta_d$, every coordinate is at least $1/(2d)$, and thus
$\KL(p\Vert q)$ is bounded by
$\sum_x|p(x)-q(x)|\log(1+2d|p(x)-q(x)|)$.
For $a\in\mathbb R$ and $z\in\mathbb R^d$, set
\begin{equation}
h(a)=|a|\log(1+2d|a|),\qquad
H(z)=\sum_xh(z(x)).
\label{eq:mw-H-definition}
\end{equation}
We measure deviations from $e_s$ in this function and take
\begin{equation}
P_s(p)=8H(p-e_s)^2.
\label{eq:mw-empirical-penalty}
\end{equation}
Since $h$ is convex and $h(2a)\leq4h(a)$, we have $H(p-q)^2\leq 8H(p-e_s)^2+8H(q-e_s)^2$, while at $\widetilde D=(D+u)/2$ the difference $\widetilde D-e_S=-(\widehat D_S-D)/2$ satisfies $\E\|\widetilde D-e_S\|_2^4=O(1/n^2)$ (\Cref{fact:empirical-l4}), which bounds $\E P_S(\widetilde D)$ by $O(\log^2(1+d/n))$.
The following lemma verifies the conditions on $P_s$ in (H3).

\begin{lemma}[Empirical moment bound]
\label[lemma]{lem:mw-empirical-moments}
The penalty in \eqref{eq:mw-empirical-penalty} is nonnegative, continuously
differentiable, and convex, vanishes at $e_s$, and satisfies
\begin{equation}
\KL(p\Vert q)^2
\leq P_s(p)+P_s(q),
\qquad p,q\in\widetilde\Delta_d.
\label{eq:mw-H-compatibility}
\end{equation}
For every $D\in\Delta_d$ and $S\sim D^n$,
\[
\E P_S(\widetilde D)\leq8\log^2(1+d/n).
\]
\end{lemma}

\begin{proof}[Proof of \Cref{thm:mirror-descent-upper-bound}]
Let $\rho=\log(1+d/n)$, $\gamma=\sqrt{2d}/n$, $R=2\sqrt2\rho$, $T=\lceil n/\sqrt d\rceil\geq1/(2\gamma)$, $\lambda=\gamma/(2R)$, $\eta=\sqrt{\gamma R}$, and $\tau=\min\{1,\gamma R\}$; these parameters depend only on $n$ and $d$. For any $D\in\Delta_d$, we apply \Cref{thm:geometric-amplification} with comparator $\theta=\widetilde D=(D+u)/2$. By \Cref{lem:efficient-amplification}, (H1) holds with this $\gamma$. For
(H2), the feedback $(f_s(Y)+\E_u[f_s])/2$ lies in $[0,1]$ and, for $Y\sim D$, has
expectation $\E_{\widetilde D}[f_s]$. By \Cref{lem:mw-empirical-moments}, (H3)
holds with this $R$. Computing each update to error at most $\tau$ in \eqref{eq:penalized-mw-approx-opt} via \Cref{prop:penalized-mw-computation} gives (H4) by \Cref{cor:penalized-kl-projection}. By \Cref{thm:geometric-amplification}, the answer $\E_r[f_S]$ of the final distribution $r$ estimates $\E_{\widetilde D}[f_S]$ with mean squared error $O(\sqrt{\gamma R})=O(d^{1/4}\sqrt{\rho/n})$. By the answer
conversion \eqref{eq:mw-mixed-answer}, the returned value
$2\E_r[f_S]-\E_u[f_S]$ has four times this error, which is
$O(d^{1/4}\sqrt{\rho/n})$ uniformly over $D\in\Delta_d$, as claimed. Every answer lies in $[0,1]$.

For the running time, since $\rho\geq \log(1+1/n)\geq 1/(2n)$, the parameters $\eta$,
$\lambda$, and $1/\tau$ are bounded by polynomials in $nd$.
\Cref{prop:penalized-mw-computation} then computes each update
using $O(d\log^3(nd))$ additional operations. Including the $O(n)$ work per
amplification proves the bound on running time. There are at most $T$
mapping evaluations, as in the simple estimator.
\end{proof}

To prove \Cref{lem:mw-empirical-moments}, we first symmetrize relative
entropy and then bound moments of $\widehat D_S-D$.

\begin{proof}[Proof of \Cref{lem:mw-empirical-moments}]
For $p,q\in\widetilde\Delta_d$,
\[
\KL(p\Vert q)\leq\KL(p\Vert q)+\KL(q\Vert p)
=\sum_x(p(x)-q(x))\log\frac{p(x)}{q(x)}
\leq H(p-q).
\]
The last inequality uses $p(x),q(x)\geq1/(2d)$. On $[0,\infty)$,
$h$ is increasing and convex, and $h(2a)\leq4h(a)$. Therefore,
$h(a+b)\leq2h(a)+2h(b)$ for $a,b\geq0$. Applying this coordinatewise, we have
\[
\KL(p\Vert q)\leq H(p-q)
\leq2H(p-e_s)+2H(q-e_s).
\]
Squaring proves \eqref{eq:mw-H-compatibility}. Since $h$ is nonnegative,
continuously differentiable, and convex, $P_s$ has these properties and
vanishes at $e_s$.

Let $z=\widehat D_S-D$, so that $\widetilde D-e_S=-z/2$ and
$\sum_x|z(x)|\leq2$. Jensen's inequality for
$a\mapsto\log(1+da)$, with weights $|z(x)|/2$ and the remaining
weight at zero, gives
\begin{equation}
H(z/2)=\sum_x\frac{|z(x)|}{2}\log(1+d|z(x)|)
\leq\log\left(1+\frac d2\|z\|_2^2\right).
\label{eq:mw-H-euclidean}
\end{equation}
To bound $\E P_S(\widetilde D)$, note that the increasing function
$\omega(a)=\log^2(1+\sqrt a)$ is concave on $[0,\infty)$.
Indeed, for $v=\sqrt a$, its derivative is
$\log(1+v)/(v(1+v))$, which decreases with $v$ because
$(1+2v)\log(1+v)\geq v$.
Using \Cref{fact:empirical-l4} in \eqref{eq:mw-H-euclidean}, we obtain
\[
\E H(z/2)^2
\leq\omega\!\left(\frac{d^2}{4}\E\|z\|_2^4\right)
\leq\omega\!\left(\frac{3d^2}{4n^2}\right)
\leq\log^2(1+d/n).
\]
Multiplying by $8$ proves the bound on $\E P_S(\widetilde D)$.
\end{proof}
\section{Gaussian Distributions}
\label{sec:gaussian-estimation}

We apply the framework of \Cref{sec:mirror-descent-amplification} to
Gaussian distributions. As described in the overview (\Cref{sec:overview}), we maintain a mixture of
Gaussian distributions, apply MW to its density, and project back onto
mixtures with a penalty around the empirical mean. This estimator is
\Cref{alg:general-amplification} for Gaussian mixtures, and
\Cref{thm:geometric-amplification} bounds its error (\Cref{sec:gaussian-mixtures}).
The guarantee is information-theoretic; we do not give an efficient
implementation of the penalized projection.
For larger sample sizes, we also give an efficient estimator
using stochastic gradient descent on the mean (\Cref{sec:gaussian-sgd}). It
uses the same chain of amplified samples with projected stochastic gradient
steps.

We prove the following formal version of \Cref{thm:gaussian-informal}.

\begin{theorem}[Gaussian distributions]
\label[theorem]{thm:gaussian-upper-bound}
Let $d,n\geq1$, and let $(f_s)_{s\in(\mathbb R^d)^n}$ be a family of
queries $f_s\colon\mathbb R^d\to[0,1]$ such that $(s,x)\mapsto f_s(x)$
is jointly measurable. There is an estimator $\phat$ that may depend on
this family and satisfies
\[
\sup_{\mu\in\mathbb R^d}
\E_{S\sim\mathcal N(\mu,I_d)^n}
\left[\left(\phat(S)-\E_{X\sim\mathcal N(\mu,I_d)}[f_S(X)]\right)^2\right]
=O\bigl(\min\{1,d^{3/4}/n\}\bigr).
\]
\end{theorem}

\paragraph{Gaussian amplification.}
To apply \Cref{thm:geometric-amplification}, we first need an amplifier.
Let $D_\mu=\mathcal N(\mu,I_d)$ and
$\widehat\mu_s=n^{-1}\sum_{i=1}^n s_i$. We use the Gaussian amplifier of
\cite[Examples~4.1 and~4.2]{axelrod2024statistical}, which preserves the
empirical mean and generates new Gaussian residuals. These residuals
have the exact distribution of $n+1$ fresh Gaussian observations after
subtracting their mean. The discrepancy is in the mean, whose covariance
is $I_d/n$ instead of $I_d/(n+1)$. Their analysis bounds total variation
distance. We calculate the $\chi^2$ divergence in the direction
required by \Cref{lem:amplification-transfer} in
\Cref{sec:gaussian-amplifier-proof}.

\begin{lemma}[Gaussian amplification]
\label[lemma]{lem:gaussian-amplification}
Given $s$, draw $Z_1,\ldots,Z_{n+1}$ independently from
$\mathcal N(0,I_d)$, set $\bar Z=(n+1)^{-1}\sum_i Z_i$, and return
\[
W_i=\widehat\mu_s+Z_i-\bar Z,\quad i=1,\ldots,n+1,
\qquad (s',y)=((W_1,\ldots,W_n),W_{n+1}).
\]
For every $\mu$ and $n\geq2$,
\begin{equation}
\chi^2\!\left(\Amp(D_\mu^n)\,
\middle\Vert\,D_\mu^{n+1}\right)
=(1-n^{-2})^{-d/2}-1\mper
\label{eq:gaussian-amplification-chi-square}
\end{equation}
If also $d\leq n^2$, this is at most $d/n^2$. The amplifier uses
$O(nd)$ arithmetic operations and independent standard normal draws.
\end{lemma}

\subsection{Multiplicative Weights over Gaussian Mixtures}
\label{sec:gaussian-mixtures}

The amplifier gives condition (H1) of \Cref{thm:geometric-amplification}. To satisfy its remaining conditions, we replace a single mean with a mixture: the squared loss of $\E_{\mathcal N(v,I_d)}[f_s]$ can be nonconvex in $v$, whereas the query expectation of a mixture of Gaussians $\mathcal N(v,I_d)$ is linear in its mixing measure, and these mixtures form a convex class containing every $D_\mu$.

Let $\mathcal P_4$ consist of the probability measures on $\mathbb R^d$
with finite fourth moment. For $\rho,\nu\in\mathcal P_4$, we write
\[
G_\nu=\int\mathcal N(v,I_d)\,\nu(dv),\qquad
B(\rho,\nu)=\KL(G_\rho\Vert G_\nu).
\]
We answer $f_s$ by $\E_{G_\nu}[f_s]$. In particular, $G_{\delta_\mu}=D_\mu$,
where $\delta_\mu$ is the point mass at $\mu$. Gaussian convolution is
injective (\Cref{app:gaussian-mixture-existence}), and thus a mixture determines
its mixing measure, and we identify the two.
In order to apply \Cref{thm:geometric-amplification}, we need a penalty
that bounds $B(\rho,\nu)^2$ and has small expectation at $\delta_\mu$.
By \Cref{cor:gaussian-mixture-kl}, $B(\rho,\nu)$ is bounded by
$\E\|V-W\|_2^2/2$ for independent $V\sim\rho$ and $W\sim\nu$.
Its square is therefore bounded by fourth moments of the mixing
measures around a common center. We use the empirical mean as this
center, since $\widehat\mu_S-\mu\sim\mathcal N(0,I_d/n)$ gives the required
moment bound at the true distribution. We set
\begin{equation}
P_s(\nu)=2\int\|v-\widehat\mu_s\|_2^4\,\nu(dv),\qquad
q_0(s)=\delta_{\widehat\mu_s}.
\label{eq:gaussian-mixture-penalty}
\end{equation}
The point mass $q_0(s)$ minimizes $P_s$ with value zero, and its mixture
is $\mathcal N(\widehat\mu_s,I_d)$. Every $G_\nu$ has a positive density, the
mixture class contains $D_\mu$, and the quartic penalty bounds the
second moment of relative entropy, allowing us to apply \Cref{thm:geometric-amplification} directly with comparator $D_\mu$.

For $\eta,\lambda>0$, we perform a multiplicative weights update on
$G_\nu$ and project back onto Gaussian mixtures with this penalty.
Writing $G_\nu(x)$ for the density, we define
\begin{equation}
\begin{aligned}
z(x)&\propto G_\nu(x)
\exp\bigl\{\eta\bigl(f_s(y)-\E_{G_\nu}[f_s]\bigr)f_s(x)\bigr\},\\
\mathsf U_s(\nu,y)
&=\argmin_{\rho\in\mathcal P_4}
\bigl\{\KL(G_\rho\Vert z)+\lambda P_s(\rho)\bigr\}.
\end{aligned}
\label{eq:gaussian-mixture-update}
\end{equation}
This is the update \eqref{eq:penalized-mw-update} of
\Cref{alg:general-amplification} for the family of Gaussian mixtures, with
the feedback $b_s(y)=f_s(y)$.
Our next lemma verifies the remaining conditions of
\Cref{thm:geometric-amplification}. We prove it after the proof of
\Cref{thm:gaussian-upper-bound}, deferring the existence, uniqueness, and
measurability of the update and the justification of its first variation
to \Cref{app:gaussian-mixture-existence}.

\begin{lemma}[Gaussian mixture update]
\label[lemma]{lem:gaussian-mixture-update}
Suppose $f_s(x)\in[0,1]$ and $(s,x)\mapsto f_s(x)$ is jointly measurable.
For every $\eta,\lambda>0$, \eqref{eq:gaussian-mixture-update} has a unique
minimizer $\nu^+$ in $\mathcal P_4$, measurable in $(s,\nu,y)$ (with the Borel $\sigma$-algebra of the weak topology on $\mathcal P_4$; \Cref{app:gaussian-mixture-existence}). For every $\rho\in\mathcal P_4$, the minimizer satisfies
\begin{equation}
\KL(G_\rho\Vert G_{\nu^+})
+\KL(G_{\nu^+}\Vert z)+\lambda P_s(\nu^+)
\leq\KL(G_\rho\Vert z)+\lambda P_s(\rho).
\label{eq:gaussian-mixture-projection}
\end{equation}
For all $\rho,\nu\in\mathcal P_4$,
\begin{equation}
B(\rho,\nu)^2\leq P_s(\rho)+P_s(\nu).
\label{eq:gaussian-mixture-compatibility}
\end{equation}
For every $\mu\in\mathbb R^d$ and $S\sim D_\mu^n$,
\begin{equation}
\E P_S(\delta_\mu)=\frac{2d(d+2)}{n^2}.
\label{eq:gaussian-mixture-penalty-moment}
\end{equation}
\end{lemma}

\begin{proof}[Proof of \Cref{thm:gaussian-upper-bound}]
If $n<\max\{2,\sqrt d\}$, the constant estimator $\phat(S)=1/2$ has mean squared error at most $1/4$, while $\min\{1,d^{3/4}/n\}=1$. Suppose now that $n\geq\max\{2,\sqrt d\}$, and run \Cref{alg:general-amplification} on $\mathcal K=\{G_\nu\mid\nu\in\mathcal P_4\}$ using the update \eqref{eq:gaussian-mixture-update}, the penalty and initialization in \eqref{eq:gaussian-mixture-penalty}, and parameters $\gamma=\sqrt d/n$, $R=\sqrt{2d(d+2)}/n$, $\lambda=\gamma/(2R)$, $T=\lceil1/\gamma\rceil$, and $\eta=\sqrt{\gamma R}$. For any $\mu\in\mathbb R^d$, we apply \Cref{thm:geometric-amplification} to $D=D_\mu$ and comparator $\theta=D_\mu$ with $\tau=0$.
\Cref{lem:gaussian-amplification} gives (H1) with this $\gamma$.
Since $f_s$ takes values in $[0,1]$ and $\theta=D_\mu$, the feedback $b_s(y)=f_s(y)$ satisfies (H2).
Since $P_s(q_0(s))=0$, \Cref{lem:gaussian-mixture-update} gives (H3) with
this $R$
and (H4) with $\tau=0$. By \Cref{thm:geometric-amplification}, the
error of its answer $\E_{G_\nu}[f_S]$ is bounded uniformly over $\mu\in\mathbb R^d$ by
$O(\min\{1,\sqrt{\gamma R}\})=O(\min\{1,d^{3/4}/n\})$.
The expectation above includes the amplifier's randomness and the choice
of $\mathcal T$. Averaging the answer over this randomness conditional on
$S$ cannot increase squared loss, and thus a deterministic estimator also
satisfies the stated bound.
\end{proof}

\begin{proof}[Proof of \Cref{lem:gaussian-mixture-update}]
To prove \eqref{eq:gaussian-mixture-compatibility}, let $V\sim\rho$ and
$W\sim\nu$ independently. By \Cref{cor:gaussian-mixture-kl}, Jensen's
inequality, and
$\|v-w\|_2^4\leq8\|v-\widehat\mu_s\|_2^4+8\|w-\widehat\mu_s\|_2^4$,
\[
B(\rho,\nu)^2
\leq\frac14\E\|V-W\|_2^4
\leq P_s(\rho)+P_s(\nu).
\]
Since $\widehat\mu_S-\mu\sim\mathcal N(0,I_d/n)$,
\[
\E P_S(\delta_\mu)=2\E\|\mu-\widehat\mu_S\|_2^4
=\frac{2d(d+2)}{n^2},
\]
proving \eqref{eq:gaussian-mixture-penalty-moment}.

Next, let $\nu^+$ be the minimizer in \eqref{eq:gaussian-mixture-update}.
Since $P_s$ is linear in the mixing measure, optimality along
$(1-\varepsilon)\nu^++\varepsilon\rho$ implies
\[
\int\bigl(G_\rho-G_{\nu^+}\bigr)
\log\frac{G_{\nu^+}}{z}\,dx
+\lambda\bigl(P_s(\rho)-P_s(\nu^+)\bigr)\geq0.
\]
Expanding relative entropy in this inequality, we obtain
\eqref{eq:gaussian-mixture-projection}. We justify the first variation and
this expansion in \Cref{app:gaussian-mixture-existence}.
\end{proof}

\subsection{Efficient Estimation by Stochastic Gradient Descent}
\label{sec:gaussian-sgd}

We now give an efficient estimator by applying stochastic gradient descent
to the mean. Its analysis follows the amplification argument above, with
an additional error from the nonlinearity of the query answer. For large
enough samples, this error is smaller than the bound in
\Cref{thm:gaussian-upper-bound}.

\begin{theorem}[Efficient Gaussian estimator]
\label[theorem]{thm:gaussian-sgd}
Let $d,n\geq1$. For every jointly measurable family
$(f_s)$ of queries $f_s\colon\mathbb R^d\to[0,1]$, there is a randomized
estimator $\phat$ satisfying
\[
\sup_{\mu\in\mathbb R^d}\E_{S\sim D_\mu^n}
\left[\left(\phat(S)-\E_{D_\mu}[f_S]\right)^2\right]
=O\left(\min\left\{1,
\frac{d^{1/4}\sqrt{d+\log n}}{n}
+\frac{(d+\log n)^2}{n^2}\right\}\right).
\]
The estimator uses $O(n\sqrt d)$ evaluations of $(s,x)\mapsto f_s(x)$
and $O(n^2\sqrt d+nd^{3/2})$ additional operations and standard normal draws.
\end{theorem}

For $d\geq\log n$, this bound is $O(d^{3/4}/n+d^2/n^2)$. It matches
\Cref{thm:gaussian-upper-bound} when $n\geq d^{5/4}$.

\paragraph{Overview.}
We use the chain of amplified samples and the reverse order of updates of
\Cref{alg:general-amplification}, with initialization $q_0(s)=\widehat\mu_s$, answer $\psi_s(v)=\E_{D_v}[f_s]$, and potential
$B(\mu,v)=\KL(D_\mu\Vert D_v)=\|v-\mu\|^2/2$.
The state is now a single mean $v\in\mathbb R^d$ rather than a mixing
measure, and thus each update is a stochastic gradient step on the squared loss
$\tfrac12(\psi_s(v)-f_s(y))^2$. For a fresh observation $Y\sim D_\mu$, the
expected gradient of this loss is the gradient of the squared estimation
error $F_s(v)=\tfrac12(\psi_s(v)-\psi_s(\mu))^2$. This error may be nonconvex
in $v$; thus the analysis of MW updates in \Cref{thm:geometric-amplification}
does not apply directly.
Since the potential is the distance to $\mu$, it suffices that $F_s$ is
approximately star-convex around its minimizer $\mu$. Indeed, since
$\psi_s$ is a Gaussian smoothing of $f_s$, its Hessian is bounded, and a
Taylor expansion of $\psi_s$ around $v$ gives
(\Cref{clm:gaussian-sgd-gradient})
\[
\langle\nabla F_s(v),v-\mu\rangle\geq F_s(v)-\frac{\|v-\mu\|^4}{8}.
\]
This inequality holds with the same constant for every query, and therefore
survives the change of query between updates. For a step size $\eta>0$,
it implies that a stochastic gradient step decreases the expected
potential by $\eta/2$ times the squared error, up to
$O(\eta^2+\eta\|v-\mu\|^4)$. Unlike the convex setting of \Cref{thm:geometric-amplification}, where the penalty only needs to control the post-update second moment of the potential, this remainder depends on the fourth moment at the \emph{pre-update} iterate (before observing the fresh sample), which a penalty in the current update does not control. Projecting each iterate onto a Euclidean ball around the empirical mean controls both the pre-update fourth moment and the post-update second moment simultaneously.

\paragraph{The estimator.}
For $n\geq2$ and $d\leq n^2$, let
\[
\gamma=\frac{\sqrt d}{n},\qquad R=\frac{d+3\log n}{n},\qquad
\kappa^2=\frac{2(d+3\log n)}{n+1}.
\]
Here $\gamma^2$ bounds the amplifier's $\chi^2$ divergence (\Cref{lem:gaussian-amplification}), and $14R^2$ bounds the second moment of the potential after an update (\Cref{clm:gaussian-sgd-moments}).
The mean of $n+1$ independent
observations from $D_\mu$ lies at distance about $\sqrt{d/(n+1)}$ from
$\mu$, and $\kappa$ exceeds this distance by a constant factor and a
logarithmic term. A ball of radius $\kappa$ around this mean therefore
contains $\mu$ except with probability at most $n^{-2}$.

The center of the ball must be computable from the pair to which we
apply the amplification guarantee. Indeed, the analysis compares an
update applied to an amplified pair $(s',y)$ with the same update
applied to $n+1$ independent observations, and thus the update may use only
$(s',y)$ and independent randomness. Since the amplifier preserves the
mean of its outputs, we can recover the input mean by
\[
c(s',y)=\frac{n\widehat\mu_{s'}+y}{n+1}=\widehat\mu_s,
\qquad (s',y)=\Amp(s).
\]
This also defines the center when $(s',y)$ is replaced with independent
observations, in which case the center is the mean of $n+1$ independent
observations. Let $\Pi_{c,\kappa}$ denote Euclidean projection onto the
ball of radius $\kappa$ around $c$.

Each update takes a stochastic gradient step on the squared loss
$L_{s,y}(v)=\tfrac12(\psi_s(v)-f_s(y))^2$. By
\Cref{lem:gaussian-smoothing-derivatives}, for independent
$Z,Z'\sim\mathcal N(0,I_d)$,
\[
\nabla L_{s,y}(v)
=\bigl(\psi_s(v)-f_s(y)\bigr)\nabla \psi_s(v)
=\E\bigl[f_s(v+Z)-f_s(y)\bigr]\cdot\E\bigl[f_s(v+Z')Z'\bigr].
\]
We estimate the two factors with disjoint Gaussian draws. With $m=2d$,
we draw independent $Z_0,\ldots,Z_m\sim\mathcal N(0,I_d)$ and define
\begin{equation}
\begin{aligned}
\widehat g
&=\bigl(f_s(v+Z_0)-f_s(y)\bigr)
\frac1m\sum_{j=1}^m f_s(v+Z_j)Z_j,\\
\mathsf U_s(v,y)&=\Pi_{c(s,y),\kappa}(v-\eta\widehat g).
\end{aligned}
\label{eq:gaussian-sgd-update}
\end{equation}
Thus $\E[\widehat g]=\nabla L_{s,y}(v)$. Each update uses independent
randomness.

These updates are projected stochastic gradient steps, which are mirror descent
steps in the Euclidean geometry. Choosing the ball radius so that the constraint holds for
the true mean with high probability introduces the $\log n$ terms in
\Cref{thm:gaussian-sgd}.

As in \Cref{sec:mirror-descent-amplification}, $q_t(s)$ denotes the final mean in a complete run with $t$
updates on input $s$. These iterates remain close to the empirical mean:
\begin{equation}
\|q_t(s)-\widehat\mu_s\|\leq\kappa\qquad\text{for every $s$ and $t\geq0$.}
\label{eq:gaussian-sgd-ball}
\end{equation}
For $t=0$, this holds since $q_0(s)=\widehat\mu_s$. For $t\geq1$, the last update
in a run on $s$ is applied with the amplified pair $(S',Y)=\Amp(s)$, and thus
its output lies in the ball of radius $\kappa$ around
$c(S',Y)=\widehat\mu_s$.

\paragraph{Analysis.}
The following lemma gives the potential recurrence for these updates.
It is the analog of the recursion \eqref{eq:geometric-recursion} in
\Cref{thm:geometric-amplification}, with an extra error term
$\eta R^2$ coming from the nonlinearity of $\psi_s$.

\begin{lemma}[Gaussian stochastic gradient updates]
\label[lemma]{lem:gaussian-sgd-recursion}
Suppose $n\geq2$ and $1\leq d\leq n^2$. For $\eta>0$, $\mu\in\mathbb R^d$, and $t\geq0$, define
\[
K_t=\E_{S\sim D_\mu^n}B(\mu,q_t(S)),\qquad
\ell_t=\E_{S\sim D_\mu^n}
\bigl(\psi_S(q_t(S))-\psi_S(\mu)\bigr)^2,
\]
where expectations include the randomness of the run computing $q_t(S)$.
Then
\begin{equation}
\frac\eta2\ell_t\leq K_t-K_{t+1}
+O(\eta R^2+\eta^2+\gamma R).
\label{eq:gaussian-sgd-recursion}
\end{equation}
\end{lemma}

We first derive \Cref{thm:gaussian-sgd} from this lemma, as in the proof of
\Cref{thm:geometric-amplification}, and then prove the lemma.

\begin{proof}[Proof of \Cref{thm:gaussian-sgd}]
If $n=1$ or $d>n^2$, the constant estimator $\phat(S)=1/2$ has mean squared error at most $1/4$ using zero evaluations, while the stated upper bound is $\Omega(1)$. Suppose now that $n\geq2$ and $d\leq n^2$, and set $T=\lceil1/\gamma\rceil$ and $\eta=\sqrt{\gamma R}$.
The proof has two steps. We first bound the error of the final mean by
summing the recurrence, and then account for estimating its answer
by sampling.

Summing \eqref{eq:gaussian-sgd-recursion} over $t=0,\ldots,T-1$, the
potentials telescope. Since $K_T\geq0$ and
$K_0=\E\|\widehat\mu_S-\mu\|^2/2=d/(2n)$, dividing by $\eta T/2$ gives
\[
\frac1T\sum_{t<T}\ell_t
\mathrel{=}O\!\left(\frac{d}{n\eta T}+R^2+\eta+\frac{\gamma R}{\eta}\right).
\]
We bound each term by $\sqrt{\gamma R}$, except for $R^2$. Since
$1/T\leq\gamma$ and $R\geq d/n$, the first term is at most
$d\gamma/(n\eta)\leq\gamma R/\eta$. With $\eta=\sqrt{\gamma R}$, both
$\eta$ and $\gamma R/\eta$ equal $\sqrt{\gamma R}$. Since the number of
updates $\mathcal T$ is chosen independently and uniformly from
$\{0,\ldots,T-1\}$, the left-hand side is the error of the answer
$\psi_S(q_{\mathcal T}(S))$, giving
\[
\E\bigl(\psi_S(q_{\mathcal T}(S))-\psi_S(\mu)\bigr)^2
\mathrel{=}O\bigl(R^2+\sqrt{\gamma R}\bigr).
\]
Since we cannot compute $\psi_S(q_{\mathcal T}(S))$ exactly, we estimate it by averaging $f_S$ over $m'=\lceil1/\sqrt{\gamma R}\rceil$ independent draws
from $D_{q_{\mathcal T}(S)}$. Conditional on $S$ and the final mean,
this average is an unbiased estimate of $\psi_S(q_{\mathcal T}(S))$ with
variance at most $1/(4m')\mathrel{\leq}\sqrt{\gamma R}/4$. The mean squared error
of $\phat(S)$ is the sum of this variance and the error above, and is
therefore $O(R^2+\sqrt{\gamma R})$. Substituting
$\sqrt{\gamma R}=d^{1/4}\sqrt{d+3\log n}/n$ gives the stated bound.
Both $\phat(S)$ and $\psi_S(\mu)$ lie in $[0,1]$, and thus the error is also at
most one.

Finally, we count operations. Since $d\leq n^2$, we have
$T=O(n/\sqrt d)$. Each amplification uses $O(nd)$ operations and
standard normal draws. Each update uses $m+2=O(d)$ query evaluations,
$O(d^2)$ standard normal draws, and $O(nd+d^2)$ operations, including
computing the center. Over $T$ updates, this gives $O(n\sqrt d)$ query
evaluations and $O(n^2\sqrt d+nd^{3/2})$ operations. Since
$\sqrt{\gamma R}\geq d^{3/4}/n$, the final average uses
$m'\leq1+n/d^{3/4}$ evaluations and $O(m'd)$ draws, which are within these
bounds.
\end{proof}

The proof of \Cref{lem:gaussian-sgd-recursion} follows the two-step
structure of the proof of \Cref{thm:geometric-amplification}. We first
analyze the last update when it is applied to $n+1$ independent
observations instead of an amplified pair. Two auxiliary claims enter
this analysis. The first collects properties of the Gaussian smoothing
$\psi_s$ and of the gradient estimate, and the second bounds fourth moments
of the iterates. We state these claims, prove the lemma, and then prove
\Cref{clm:gaussian-sgd-gradient}. We prove \Cref{clm:gaussian-sgd-moments} in
\Cref{app:gaussian-sgd-moments}.

\begin{claim}[Gaussian smoothing]
\label[claim]{clm:gaussian-sgd-gradient}
Let $s\in(\mathbb R^d)^n$ and $v,\mu\in\mathbb R^d$, and let
$F_s(v)=\tfrac12(\psi_s(v)-\psi_s(\mu))^2$. Let $\widehat g$ be defined by
\eqref{eq:gaussian-sgd-update} with $y=Y\sim D_\mu$, independent of
$Z_0,\ldots,Z_m$. Then $\E\widehat g=\nabla F_s(v)$,
\[
\langle\nabla F_s(v),v-\mu\rangle
\geq F_s(v)-\frac{\|v-\mu\|^4}{8},
\qquad
\E\|\widehat g\|^2\leq2.
\]
\end{claim}

\begin{claim}[Fourth moments]
\label[claim]{clm:gaussian-sgd-moments}
Let $(\bar S,\bar Y)\sim D_\mu^{n+1}$, and let $v=q_t(\bar S)$,
$V=\mathsf U_{\bar S}(v,\bar Y)$, and $\bar c=c(\bar S,\bar Y)$. Then
\[
\E\|v-\mu\|^4\leq56R^2,\qquad
\E\|V-\mu\|^4\leq56R^2,\qquad
\Prb(\|\bar c-\mu\|>\kappa)\leq n^{-2}.
\]
\end{claim}

\begin{proof}[Proof of \Cref{lem:gaussian-sgd-recursion}]
Let $(\bar S,\bar Y)$, $v$, $V$, and $\bar c$ be as in
\Cref{clm:gaussian-sgd-moments}. We prove that
\begin{align}
\E B(\mu,V)&\leq K_t-\frac\eta2\ell_t
+7\eta R^2+\eta^2+4\gamma R,
\label{eq:gaussian-sgd-progress}\\
K_{t+1}&\leq\E B(\mu,V)+4\gamma R.
\label{eq:gaussian-sgd-transfer}
\end{align}
Combining the two inequalities proves
\eqref{eq:gaussian-sgd-recursion}. The first bounds the progress of the
last update when it uses independent observations, and the second
accounts for replacing them with an amplified pair.

\emph{Transfer to the amplified pair.}
We first prove \eqref{eq:gaussian-sgd-transfer}, assuming
\Cref{clm:gaussian-sgd-moments}. By \Cref{lem:gaussian-amplification}, the
amplifier's $\chi^2$ divergence is at most $d/n^2=\gamma^2$. Applying
\Cref{lem:amplification-transfer} with $G=B(\mu,\cdot)$,
and then \Cref{clm:gaussian-sgd-moments}, we obtain
\[
K_{t+1}\leq\E B(\mu,V)
+\gamma\sqrt{\E B(\mu,V)^2}
=\E B(\mu,V)+\frac{\gamma}2\sqrt{\E\|V-\mu\|^4}
\leq\E B(\mu,V)+4\gamma R.
\]

\emph{Progress with independent observations.}
It remains to prove \eqref{eq:gaussian-sgd-progress}. Let
$v'=v-\eta\widehat g$ be the iterate before projection, so that
$V=\Pi_{\bar c,\kappa}(v')$, and let $E$ be the event
$\|\bar c-\mu\|>\kappa$. We first remove the projection. Outside $E$, the
ball around $\bar c$ contains $\mu$, and Euclidean projection onto a
convex set does not increase the distance to any of its points, giving
$\|V-\mu\|\leq\|v'-\mu\|$. On $E$, we use only that $V$ lies in the ball:
$\|V-\mu\|\leq\kappa+\|\bar c-\mu\|\leq2\|\bar c-\mu\|$. Thus
\begin{equation}
B(\mu,V)\leq\frac12\|v'-\mu\|^2+2\|\bar c-\mu\|^2\1_E.
\label{eq:gaussian-sgd-projection}
\end{equation}

Next, we bound the expectation of the first term. Expanding the square, we have
\[
\frac12\|v'-\mu\|^2
=B(\mu,v)-\eta\langle\widehat g,v-\mu\rangle
+\frac{\eta^2}2\|\widehat g\|^2.
\]
Conditional on $\bar S$ and the randomness used to compute $v$, the
observation $\bar Y$ has distribution $D_\mu$ and is independent of the
Gaussian draws in the update. We may therefore apply
\Cref{clm:gaussian-sgd-gradient} with $s=\bar S$, which gives
$\E[\langle\widehat g,v-\mu\rangle\mid\bar S,v]
\geq F_{\bar S}(v)-\|v-\mu\|^4/8$, and then take expectations. Since
$\E F_{\bar S}(v)=\ell_t/2$ and
$\E B(\mu,v)=K_t$, \Cref{clm:gaussian-sgd-moments} gives
\[
\frac12\E\|v'-\mu\|^2
\leq K_t-\frac\eta2\ell_t+\frac\eta8\E\|v-\mu\|^4+\eta^2
\leq K_t-\frac\eta2\ell_t+7\eta R^2+\eta^2.
\]

Finally, we bound the second term in
\eqref{eq:gaussian-sgd-projection}. The center
$\bar c$ is the mean of $n+1$ independent observations from $D_\mu$; thus
$\bar c-\mu\sim\mathcal N(0,I_d/(n+1))$ and
$\E\|\bar c-\mu\|^4=d(d+2)/(n+1)^2$. By Cauchy--Schwarz and
\Cref{clm:gaussian-sgd-moments},
\[
2\E\bigl[\|\bar c-\mu\|^2\1_E\bigr]
\leq2\sqrt{\E\|\bar c-\mu\|^4\,\Prb(E)}
\leq\frac{2\sqrt{d(d+2)}}{n(n+1)}
\leq\frac{2\sqrt3\,d}{n^2}
\leq4\gamma R,
\]
where the last step uses $\gamma R\geq d^{3/2}/n^2\geq d/n^2$.
Taking expectations in \eqref{eq:gaussian-sgd-projection} and combining
the last two displays proves \eqref{eq:gaussian-sgd-progress}.
\end{proof}

\begin{proof}[Proof of \Cref{clm:gaussian-sgd-gradient}]
By \Cref{lem:gaussian-smoothing-derivatives}, $\psi_s$ is twice
continuously differentiable, with $\|\nabla \psi_s(v)\|\leq1$ and
$\|\nabla^2\psi_s(v)\|_{\mathrm{op}}\leq1$ for every $v$.

We write $\widehat g=Ab$, where
$A=f_s(v+Z_0)-f_s(Y)$ and
$b=m^{-1}\sum_{j=1}^mf_s(v+Z_j)Z_j$. These factors are independent,
$\E A=\psi_s(v)-\psi_s(\mu)$, and $\E b=\nabla \psi_s(v)$. Therefore
$\E\widehat g=(\psi_s(v)-\psi_s(\mu))\nabla \psi_s(v)=\nabla F_s(v)$.
For the star-convexity bound, Taylor's theorem and the Hessian bound give
$\psi_s(\mu)=\psi_s(v)-\langle\nabla \psi_s(v),v-\mu\rangle+r'$ with
$|r'|\leq\|v-\mu\|^2/2$. Therefore,
\[
\langle\nabla F_s(v),v-\mu\rangle
=(\psi_s(v)-\psi_s(\mu))^2+(\psi_s(v)-\psi_s(\mu))r'
=2F_s(v)+(\psi_s(v)-\psi_s(\mu))r'.
\]
Since $|(\psi_s(v)-\psi_s(\mu))r'|\leq F_s(v)+r'^2/2\leq
F_s(v)+\|v-\mu\|^4/8$, this proves the bound.
For the second moment, $|A|\leq1$, and the summands of $b$ are
independent with mean $\nabla \psi_s(v)$. Thus
\[
\E\|\widehat g\|^2\leq\E\|b\|^2
\leq\|\nabla \psi_s(v)\|^2+\frac1m\E\|Z\|^2
\leq1+\frac dm\leq2.
\qedhere
\]
\end{proof}
\section{Smoothed Sample Size}
\label{sec:random-horizon}

With a randomly chosen sample size, the observations themselves give the
fresh feedback required by the MW progress bound. Each next observation
is independent of both the query and the estimate computed from the
preceding observations. In an independent sample of size $t+1$, the prefix of length $t$ and the final observation have the distribution $D^{t+1}$ that amplification only approximates at a fixed sample size. Because there is no amplification error, we can telescope relative entropy along the prefixes of one sample without a chain of amplified samples, a projection, or a penalty.
Choosing the sample size uniformly turns the average squared error along
these updates into the error of our estimator.

\begin{theorem}[Smoothed sample size]
\label[theorem]{thm:random-horizon-upper-bound}
Let $n,h\geq1$ and $d\geq2$ be integers, and let $X_1,X_2,\ldots$ be
independent observations from $D\in\Delta_d$. Independently of the
observations, choose $\tau$ uniformly from $\{n,\ldots,n+h-1\}$.
Let $\dde$ be a known mapping defined on samples of every size in this
interval, and write $S_t=(X_1,\ldots,X_t)$.
There is an estimator $\phat$ based on $(\tau,S_\tau)$ and $\dde$ such that
\[
\sup_{D\in\Delta_d}
\E_{\tau,S_\tau}
\left[\left(\phat(\tau,S_\tau)-\E_D[\dde(S_\tau)]\right)^2\right]
=
O\left(\min\left\{1,
\sqrt{\frac{\log(1+d/n)}{h}}
\right\}\right).
\]
It uses at most $h$ evaluations of $\dde$ and $O(n+hd)$ additional
operations.
\end{theorem}

Taking $h=n$ gives \Cref{thm:smoothed-intro}, which bounds the error by
$O(\sqrt{\log(1+d/n)/n})$ using at most $2n-1$ observations.
Our estimator starts from the Laplace predictor
$\Lap(s)(x)=(N_x(s)+1)/(|s|+d)$ on $S_n$, where $N_x(s)$ counts occurrences of $x$ in $s$, and we combine the MW progress inequality (\Cref{fact:mw-progress}) with the following relative entropy bound.

\begin{fact}[Laplace prediction {\cite[Proposition~4]{MourtadaGaiffas2022}}]
\label[fact]{fact:laplace-predictive-risk}
For every $D\in\Delta_d$ and $S\sim D^n$,
\[
\E\KL(D\Vert\Lap(S))
\leq\log\frac{n+d}{n+1}\leq\log(1+d/n).
\]
\end{fact}

At sample size $t$ in \Cref{alg:random-sample-size}, both the current distribution $Q_t$ and the selected
query $f_t=\dde(S_t)$ depend only on $S_t$. If another observation is
available, we use $f_t(X_{t+1})$ for the next MW update. When the sample stops at $\tau$, the algorithm outputs $\E_{Q_\tau}[f_\tau]$ after $\tau-n$ updates without needing $X_{\tau+1}$.

\begin{algorithm}[H]
\caption{Estimation with a smoothed sample size}
\label{alg:random-sample-size}
\small
\begin{algorithmic}[1]
\Statex \textbf{Input} Sample $S_\tau$, mapping $\dde$, and $n,h,d$
\State $\eta\gets\sqrt{\log(1+d/n)/h}$
\State $Q_n\gets\Lap(S_n)$
\Statex
\For{$t=n,\ldots,\tau-1$}
    \State $f_t\gets\dde(S_t)$
    \State $Q_{t+1}\gets\MW_\eta(Q_t,f_t,f_t(X_{t+1}))$
\EndFor
\State $f_\tau\gets\dde(S_\tau)$
\State \Return $\phat(\tau,S_\tau)=\E_{Q_\tau}[f_\tau]$
\end{algorithmic}
\end{algorithm}

\begin{proof}[Proof of \Cref{thm:random-horizon-upper-bound}]
\Cref{alg:random-sample-size} uses only $S_\tau$. For fixed $D\in\Delta_d$, in the analysis we
continue the updates through $Q_{n+h}$ on an independent
sequence from $D$, independent of $\tau$; this continuation does not change the
estimate at time $\tau$. For $n\leq t<n+h$, write $f_t=\dde(S_t)$, and define
\[
\begin{aligned}
K_t&=\E_{S_t}\KL(D\Vert Q_t),
&&n\leq t\leq n+h,\\
\ell_t&=\E_{S_t}\bigl(\E_{Q_t}[f_t]-\E_D[f_t]\bigr)^2,
&&n\leq t<n+h.
\end{aligned}
\]
Conditional on $S_t$, both $Q_t$ and $f_t$ are fixed,
and $X_{t+1}$ is independent of them. By the MW progress bound
(\Cref{fact:mw-progress}),
$\eta\ell_t\leq K_t-K_{t+1}+\eta^2/8$.
Averaging over $\tau$ and summing over $t=n,\ldots,n+h-1$, we obtain
\begin{align*}
\E_{\tau,S_\tau}
\bigl(\phat(\tau,S_\tau)-\E_D[\dde(S_\tau)]\bigr)^2
&=\frac1h\sum_{t=n}^{n+h-1}\ell_t\\
&\leq\frac{K_n-K_{n+h}}{\eta h}+\frac\eta8\\
&\leq\frac{\log(1+d/n)}{\eta h}+\frac\eta8.
\end{align*}
Here we used $K_{n+h}\geq0$ and the Laplace prediction bound
(\Cref{fact:laplace-predictive-risk}). Substituting $\eta$ proves the
claimed rate. The risk is also at most one.

Initialization takes $O(n+d)$ operations, and each update takes $O(d)$
additional operations. Since there are $\tau-n$ updates and one final
mapping evaluation, we use at most $h$ mapping evaluations.
Accessing each prefix without copying it keeps the total additional
work at $O(n+hd)$.
\end{proof}
\section{Lower Bounds}
\label{sec:risk-lower-bounds}

In this section, we prove \Cref{thm:risk-lower-bound}, which we restate
for convenience.

\risklowerbound*

To prove our lower bounds, we choose a prior $\Pi$ over distributions in $\Delta_d$ and construct an event mapping $\dde\colon[d]^n\to\{0,1\}^{[d]}$ whose probability under the unknown distribution has large expected posterior variance given the sample $S$. As is implicit in prior lower-bound proofs~\cite{gastpar2024which,FeldmanKRTT26}, expected posterior variance lower bounds the worst-case risk of every estimator; we include the short proof for completeness.

\begin{lemma}[Lower bound from a prior]
\label[lemma]{cor:bayes-lower-bound}
Let $P\sim\Pi$ be any prior on $\Delta_d$ and let $S\mid P\sim P^n$.
For every mapping $\dde\colon[d]^n\to[0,1]^{[d]}$ and every randomized estimator
$\phat$, we have
\[
\sup_{D\in\Delta_d}\E_{S\sim D^n}
\bigl[(\phat(S)-\E_D[\dde(S)])^2\bigr]
\geq\E_S\Var\bigl(\E_P[\dde(S)]\mid S\bigr).
\]
\end{lemma}
\begin{proof}
Worst-case risk is at least average risk under $\Pi$. Writing $Y=\E_P[\dde(S)]$, conditional on $S$ the estimator's internal randomness is independent of $P$, and thus $\E[(\phat(S)-Y)^2\mid S]=\E[(\phat(S)-\E[Y\mid S])^2\mid S]+\Var(Y\mid S)\geq\Var(Y\mid S)$. Averaging over $S$ proves the claim.
\end{proof}

We follow the basic approach
of~\cite[Theorem~2.1]{gastpar2024which}
and~\cite[Theorems~7.2 and~7.3]{FeldmanKRTT26}. With some probability $p$,
the sample leaves a candidate distribution whose posterior probability is
bounded away from zero and one. The mapping selects an event whose probability
under this candidate differs by at least $\Delta$ from its probability under
every other candidate, giving expected posterior variance $\Omega(p\Delta^2)$. Both of our constructions use the
domain $[m]\times\{-1,1\}$ with $m=\lfloor d/2\rfloor$, which we identify with $[2m]\subseteq[d]$ (assigning zero probability and query value zero to $d$ when $d$ is odd, and selecting the empty event if $S$ contains $d$). Each observation
is a uniformly random coordinate $i\in[m]$ together with a sign that
carries information about a hidden word.

For $n\leq m/8$ (\Cref{sec:small-sample-lower-bound}), let $j$ be the
most likely number of distinct coordinates among the $n$ observations.
The sign at coordinate $i$ is $(-1)^{\langle A_i,\theta\rangle}$ for a hidden
parameter $\theta\in\mathbb F_2^{j+1}$ and known rows $A_i$. Each observed
coordinate thus reveals one linear equation in $\theta$, as in the parity
construction of~\cite[Theorem~7.2]{FeldmanKRTT26}. If the observed
coordinates give $j$ independent equations, two equally likely parameters
remain, and they disagree on at least a quarter of the coordinates with constant probability, making $\Delta\geq1/4$. Here $p$ is proportional to the
probability that $j$ distinct coordinates are observed: a constant when $n=O(\sqrt m)$, and $\Omega(\sqrt m/n)$ for larger $n$ because repeated coordinates make the count of distinct coordinates fluctuate by $O(n/\sqrt m)$. Whenever two equally likely parameters with separation $1/4$ remain, every estimate errs by more than $1/16$ on at least one of them, which simultaneously yields our confidence lower bound for $d\geq16n$ (\Cref{lem:small-sample-lower-bound}).

For $n\geq m/8$ (\Cref{sec:large-sample-lower-bound}),
each sign agrees with the corresponding bit of a hidden codeword only
with probability about $(1+\beta)/2$, where $\beta=a\sqrt{m/n}$ for a small
constant $a>0$. Each coordinate is observed $n/m$ times in expectation, and thus the
sample makes the true word about $e^{O(a^2m)}$ times more
likely than a uniformly random word. If the code has about that many
words, some other codeword can then be about as likely as the true
one. The logarithm of this likelihood ratio fluctuates by $O(\sqrt m)$. We therefore choose
the size of a random code from a window of $O(\sqrt m)$ values, so that
with probability $p=\Omega(1/\sqrt m)$ the most likely codeword given the
sample has posterior probability bounded away from zero and one
(\Cref{lem:random-code-window}). The mapping selects the event aligned
with this most likely codeword. Its probabilities under two
codewords at constant relative distance differ by
$\Delta=\Omega(\beta)=\Omega(\sqrt{m/n})$, giving $p\Delta^2=\Omega(\sqrt m/n)$ and covering all remaining sample sizes.
We prove the lemmas of this section in \Cref{app:lower-bound-proofs}.

\subsection{Small Samples: Hidden Linear Equations}
\label{sec:small-sample-lower-bound}

Let $M$ be the number of distinct coordinates among $n$ independent
uniform draws from $[m]$. We first show that some value of $M$ has
probability $\Omega(\min\{1,\sqrt m/n\})$.

\begin{lemma}[Distinct coordinates]
\label[lemma]{lem:distinct-coordinates}
For every $n,m\geq1$, there is $j\in[n]$ such that
\[
\Prb(M=j)\mathrel{\geq}\frac3{20}\min\left\{1,\frac{\sqrt m}n\right\}.
\]
\end{lemma}

\Cref{con:linear-equations-lower-bound} specifies the distribution and
the selected event.

\begin{construction}[H]
\caption{Lower bound using linear equations}
\label{con:linear-equations-lower-bound}
\small
\noindent\textbf{Parameters.}
Let $n\leq m/8$. Choose $j\in[n]$ maximizing $\Prb(M=j)$, and fix rows $A_1,\ldots,A_m\in\mathbb F_2^{j+1}$ for which the event $E_A(S)$ in the proof of \Cref{lem:small-sample-lower-bound} has probability at least $\Prb(M=j)/4$.

\smallskip
\noindent\textbf{Prior and distribution.}
Draw $\theta$ uniformly from
$\mathbb F_2^{j+1}$ and observe $S\sim D_\theta^n$, where
\[
D_\theta(i,b)=\frac1m\1\{b=(-1)^{\langle A_i,\theta\rangle}\},
\qquad (i,b)\in[m]\times\{-1,1\},
\]
with inner products over $\mathbb F_2$.

\smallskip
\noindent\textbf{Selected event.}
If the equations revealed by $S$ have two solutions
$\theta_0,\theta_1$, list them in a fixed order and set
\[
\dde_A(S)(i,b)=\1\{D_{\theta_0}(i,b)>D_{\theta_1}(i,b)\}.
\]
Otherwise, select the empty event.
\end{construction}

\begin{lemma}[Small samples and confidence lower bound]
\label[lemma]{lem:small-sample-lower-bound}
\label[lemma]{cor:confidence-lower-bound}
For every $n\geq1$ and $d\geq16n$, there is a mapping
$\dde\colon[d]^n\to\{0,1\}^{[d]}$ such that every randomized estimator
$\phat$ satisfies
\[
\sup_{D\in\Delta_d}
\Prb_{S\sim D^n}\!\left(\left|\phat(S)-\E_D[\dde(S)]\right|>\frac1{16}\right)
=\Omega\!\left(\min\!\left\{1,\frac{\sqrt d}{n}\right\}\right),
\]
where the probability includes the estimator's randomness. In particular,
\[
\sup_{D\in\Delta_d}
\E_{S\sim D^n}\!\left[\left(\phat(S)-\E_D[\dde(S)]\right)^2\right]
=\Omega\!\left(\min\!\left\{1,\frac{\sqrt d}{n}\right\}\right).
\]
\end{lemma}

\subsection{Large Samples: Noisy Codewords}
\label{sec:large-sample-lower-bound}

\Cref{con:fixed-sample-code-lower-bound} specifies the distribution and
the selected event for $n\geq m/8$.

\begin{construction}[H]
\caption{Code construction for large samples}
\label{con:fixed-sample-code-lower-bound}
\small
\noindent\textbf{Parameters.}
Let $n\geq m/8$ and $\beta=a\sqrt{m/n}$ for a sufficiently small constant
$a>0$.
Fix a code $\mathcal C\subseteq\{-1,1\}^m$ with pairwise Hamming distances greater than $m/4$ chosen via \Cref{lem:random-code-window} as in the proof of \Cref{lem:large-sample-lower-bound}.

\smallskip
\noindent\textbf{Prior and distribution.}
Draw $V$ uniformly from $\mathcal C$ and observe $S\sim D_V^n$, where
\[
D_v(j,\sigma)=\frac{e^{\beta\sigma v_j}}{2m\cosh\beta},
\qquad (j,\sigma)\in[m]\times\{-1,1\}.
\]

\smallskip
\noindent\textbf{Selected event.}
Let $v(S)$ be the most likely codeword given $S$, breaking ties in a
fixed order, and set
\[
\dde(S)(j,\sigma)=\1\{\sigma=v(S)_j\}.
\]
\end{construction}

Given an observation $Y$ drawn from $p(\cdot\mid V_*)$ for a chosen codeword $V_*$ in a random list $\mathcal C$ of $1+2^\ell$ uniform words, the posterior probability of $V_*$ is $L_*/(L_*+W)$, where $L_*=p(Y\mid V_*)/p_0(Y)$ is its likelihood ratio relative to a random word and $W$ is the sum of likelihood ratios of the $2^\ell$ other words. The sum $W$ has conditional mean $2^\ell$, and its part truncated at $L_*$ is of order $2^\ell$ with constant probability whenever $2^\ell$ matches $L_*$ within constant factors, keeping the posterior probability of $V_*$ bounded away from $0$ and $1$. Since $\log L_*$ concentrates within $O(\sqrt k)$ of its mean $J$ when $\Var(\log L_*)\leq k$, sampling $\ell$ uniformly from a window of $O(\sqrt k)$ integers around $J/\log 2$ hits $\lceil\log_2 L_*\rceil$ with probability $\Omega(1/\sqrt k)$. We prove the resulting general lemma in \Cref{sec:ambiguous-code}.

\begin{lemma}[Choosing the code size]
\label[lemma]{lem:random-code-window}
For every integer $k\geq1$, let $V$ be uniform on $\{-1,1\}^k$, let $Y\sim p(\cdot\mid V)$ have conditional probability mass or density $p(y\mid V)$, and suppose that the log-likelihood ratio
\[
\Lambda=\log\frac{p(Y\mid V)}{p_0(Y)},
\qquad\text{where }p_0(y)=\E_{U\sim\mathrm{Unif}(\{-1,1\}^k)}\bigl[p(y\mid U)\bigr],
\]
satisfies $\Var(\Lambda)\leq k$ with mean $J=\E\Lambda$.
Let $\mathcal I$ be the set of integers $\ell\geq0$ with
$|\ell\log2-J|\leq6\sqrt k$. Choose $\ell$ uniformly from $\mathcal I$, form a list $\mathcal C$ of
$1+2^\ell$ independent uniform words in $\{-1,1\}^k$, choose a uniform
index in the list, and draw $Y$ from $p(\cdot\mid w)$ for the word $w$ at
this index. With probability at least $c/\sqrt k$ over $\ell$, $\mathcal C$,
the index, and $Y$, the most likely index
in $\mathcal C$ given $\mathcal C$ and $Y$ has posterior probability in $[c,1-c]$, for an absolute constant $c>0$.
\end{lemma}

\begin{lemma}[Large samples]
\label[lemma]{lem:large-sample-lower-bound}
There is an absolute constant $m_0\geq1$ such that for every $d\geq2m_0$ and $n\geq\lfloor d/2\rfloor/8$, there is a mapping $\dde\colon[d]^n\to\{0,1\}^{[d]}$ such that every randomized estimator $\phat$ satisfies
\[
\sup_{D\in\Delta_d}
\E_{S\sim D^n}\!\left[\left(\phat(S)-\E_D[\dde(S)]\right)^2\right]
=\Omega\!\left(\frac{\sqrt d}{n}\right).
\]
\end{lemma}

\subsection{Proof of Theorem~\ref{thm:risk-lower-bound}}
\label{sec:uniform-risk-lower-bound}

\begin{proof}[Proof of \Cref{thm:risk-lower-bound}]
Let $m=\lfloor d/2\rfloor$ and let $m_0$ be the constant from \Cref{lem:large-sample-lower-bound}. Suppose first that $m\geq m_0$. Since $16n$ is even, $d\geq16n$ holds if and only if $n\leq m/8$. Thus \Cref{lem:small-sample-lower-bound} proves the bound when $n\leq m/8$, and \Cref{lem:large-sample-lower-bound} proves the bound $\Omega(\sqrt d/n)=\Omega(\min\{1,\sqrt d/n\})$ when $n>m/8$.

Suppose now that $m<m_0$, so that $d<2m_0+2$ and $\min\{1,\sqrt d/n\}=\Theta(1/n)$. Consider the constant mapping $\dde(s)=\1_{\{1\}}$ and the uniform prior $\Pi$ over the two distributions $D_\pm$ supported on $\{1,2\}$ with $D_\pm(1)=1/2\pm\delta$ for $\delta=1/(4\sqrt n)$. By additivity of relative entropy over products (\Cref{fact:kl-chain-rule}), $\KL(D_+^n\Vert D_-^n)=n\cdot 2\delta\log\frac{1+2\delta}{1-2\delta}\leq 32n\delta^2/3=2/3$, and thus Pinsker's inequality~\cite[Lemma~11.6.1]{CoverThomas2006} gives $\TV(D_+^n,D_-^n)\leq\sqrt{1/3}$. Under $\Pi$, the posterior probabilities $\pi_+,\pi_-$ of $D_+,D_-$ given $S$ satisfy $\E_S[\pi_+\pi_-]\geq\frac12\E_S[\min\{\pi_+,\pi_-\}]=\frac14(1-\TV(D_+^n,D_-^n))$, giving $\E_S\Var(P(\{1\})\mid S)=4\delta^2\E_S[\pi_+\pi_-]\geq(1-1/\sqrt3)\delta^2=\Omega(1/n)$ and proving the bound by \Cref{cor:bayes-lower-bound}.
\end{proof}
\section{Open Questions}

We prove that for every mapping over a domain of size
$d=o(n^2/\log^2 n)$ the mean squared error of our estimator tends to zero. Can we remove the logarithmic factor
and allow $d=o(n^2)$, as asked in~\cite{FeldmanKRTT26}?

The bounds proved here leave a gap in the mean squared error when
$n\gg\sqrt d$. At $n=d$, they are
$\Omega(n^{-1/2})$ and $O(n^{-1/4})$. For $n\geq d$, they are
$\Omega(\sqrt d/n)$ and $O(d^{3/4}/n)$.
Determining the optimal dependence on $d$ across all sample sizes $n\gg\sqrt d$ remains open (a stronger lower bound of $\Omega(d^{2/3}/n)$ for $n=\Omega(d\log d)$, whose proof is not included in this version, is noted in the \hyperref[par:ai-methodology]{AI methodology} statement).

For Gaussian distributions $\mathcal N(\mu,I_d)$, our mixture estimator achieves mean squared error $O(\min\{1,d^{3/4}/n\})$, while our polynomial-time SGD estimator matches this bound for $n\geq d^{5/4}$ (with $d\geq\log n$). Finding a polynomial-time estimator that achieves $O(d^{3/4}/n)$ for all $n\geq d^{3/4}$ (for instance, by efficiently approximating the Gaussian mixture projection), as well as extending these bounds to Gaussians with unknown covariance and characterizing which general distribution families admit data-dependent probability estimation below the sample complexity of distribution learning, are natural directions for future work.

\appendix
\addtocontents{toc}{\protect\newpage}
\section{Additional Preliminaries}
\label{app:additional-preliminaries}
\label{app:auxiliary-facts}

In this section we collect auxiliary facts and prove statements deferred
from \Cref{sec:preliminaries}.

\subsection{Matrix Notation}

We write $\tr(A)$ for the trace,
$\operatorname{diag}(v)$ for the diagonal matrix with diagonal $v$,
and $I_r$ for the $r\times r$ identity matrix.
We write $\mathcal N(m,\Sigma)$ for the Gaussian distribution with mean
$m$ and covariance $\Sigma$.

\subsection{Empirical Estimation}
\label{app:empirical-estimation}

To analyze the penalized update, we bound the fourth moment of the
empirical distribution.

\begin{fact}[Empirical fourth moment]
\label[fact]{fact:empirical-l4}
For $D\in\Delta_d$, $n\geq1$, and $S\sim D^n$,
$\E\|\widehat D_S-D\|_2^4\leq3/n^2$.
\end{fact}

\begin{proof}
Let $Z_i=\1_{\{X_i\}}-D$ and
$\Sigma=\E[Z_iZ_i^\top]=\operatorname{diag}(D)-DD^\top$.
Since the $Z_i$ are independent and have mean zero, a term in the fourth
moment expansion vanishes whenever one of its indices occurs only once.
Therefore,
\begin{align*}
n^4\E\|\widehat D_S-D\|_2^4
&=n\E\|Z_1\|_2^4
+n(n-1)(\tr\Sigma)^2\\
&\quad+2n(n-1)\tr(\Sigma^2)
\leq2n+3n(n-1)\leq3n^2.
\end{align*}
Here $\|Z_1\|_2^2\leq2$ and $\E\|Z_1\|_2^2\leq1$, and thus
$\E\|Z_1\|_2^4\leq2$. Since $\Sigma$ is positive semidefinite with
$\tr\Sigma\leq1$, its eigenvalues satisfy
$\tr(\Sigma^2)\leq(\tr\Sigma)^2\leq1$.
\end{proof}

\subsection{Gaussian Distributions and Mixtures}

For the Gaussian amplifier, we use the scalar case of the Gaussian
divergence formula in \cite[Lemma~4]{DiakonikolasKane2021}.

\begin{fact}[Gaussian $\chi^2$ divergence]
\label[fact]{fact:gaussian-chi-square}
For $0<v<2$,
\[
1+\chi^2\bigl(\mathcal N(0,v)\Vert\mathcal N(0,1)\bigr)
=\bigl(v(2-v)\bigr)^{-1/2}.
\]
\end{fact}

We use the chain rule to bound relative entropy for Gaussian mixtures
before discarding the component index.

\begin{fact}[Chain rule for relative entropy {\cite[Theorem~2.5.3]{CoverThomas2006}}]
\label[fact]{fact:kl-chain-rule}
For joint distributions $P_{XY},Q_{XY}$,
\[
\KL(P_{XY}\Vert Q_{XY})
=\KL(P_X\Vert Q_X)
+\E_{X\sim P_X}\KL(P_{Y\mid X}\Vert Q_{Y\mid X}).
\]
In particular, discarding a coordinate cannot increase relative entropy.
Applying a fixed randomized procedure to both distributions also cannot
increase it. For product distributions, relative entropy is the sum of
the coordinatewise relative entropies.
\end{fact}

\begin{corollary}[Relative entropy between Gaussian mixtures]
\label[corollary]{cor:gaussian-mixture-kl}
For probability distributions $\rho,\nu$ on $\mathbb R^d$, let
$G_\rho=\int\mathcal N(v,I_d)\,\rho(dv)$ and
$G_\nu=\int\mathcal N(w,I_d)\,\nu(dw)$.
For every coupling $\pi$ of $\rho,\nu$,
\[
\KL(G_\rho\Vert G_\nu)
\leq\frac12\E_{(V,W)\sim\pi}\|V-W\|_2^2.
\]
\end{corollary}
\begin{proof}
Draw $(V,W)\sim\pi$, then draw $X$ from either $\mathcal N(V,I_d)$
or $\mathcal N(W,I_d)$. By the chain rule
(\Cref{fact:kl-chain-rule}), the relative entropy between these two
distributions of $(V,W,X)$ is $\E\|V-W\|_2^2/2$. Discarding $(V,W)$ leaves
$G_\rho$ and $G_\nu$, proving the claim.
\end{proof}

The stochastic gradient estimator in \Cref{sec:gaussian-sgd} uses the
following expressions for the derivatives of a Gaussian smoothing.
They are the first- and second-order forms of Stein's lemma.
If $f$ is smooth, then
$\nabla \psi(v)=\E[\nabla f(v+Z)]$, and Gaussian integration by parts
rewrites this as $\E[f(v+Z)Z]$, and similarly for the Hessian. Stein's
lemma requires $f$ to be (weakly) differentiable, while our queries are
only measurable. We therefore apply the integration by parts to the
Gaussian density instead of to $f$.

\begin{lemma}[Derivatives of Gaussian smoothing]
\label[lemma]{lem:gaussian-smoothing-derivatives}
Let $f\colon\mathbb R^d\to[0,1]$ be measurable, let
$Z\sim\mathcal N(0,I_d)$, and define $\psi(v)=\E[f(v+Z)]$. Then $\psi$ is
twice continuously differentiable, with
\[
\nabla \psi(v)=\E[f(v+Z)Z],\qquad
\nabla^2\psi(v)=\E\bigl[f(v+Z)(ZZ^\top-I_d)\bigr].
\]
Moreover, $\|\nabla \psi(v)\|\leq1$ and
$\|\nabla^2\psi(v)\|_{\mathrm{op}}\leq1$ for every $v\in\mathbb R^d$.
\end{lemma}

\begin{proof}
Let $\varphi(z)=(2\pi)^{-d/2}e^{-\|z\|^2/2}$ be the standard Gaussian
density. Substituting $x=v+z$ moves the dependence on $v$ from the
bounded function $f$ to the smooth density:
\[
\psi(v)=\int f(x)\varphi(x-v)\,dx.
\]
Since $\nabla\varphi(z)=-z\varphi(z)$ and
$\nabla^2\varphi(z)=(zz^\top-I_d)\varphi(z)$, the chain rule gives
\[
\nabla_v\varphi(x-v)=(x-v)\varphi(x-v),\qquad
\nabla_v^2\varphi(x-v)=\bigl((x-v)(x-v)^\top-I_d\bigr)\varphi(x-v).
\]

We justify differentiating under the integral. Fix $v_0$ and consider
$v$ with $\|v-v_0\|\leq1$. Then $\|x-v\|\leq\|x-v_0\|+1$ and
$\|x-v\|\geq(\|x-v_0\|-1)_+$, and thus the entries of $\varphi(x-v)$ and of its
first two derivatives in $v$ are bounded in absolute value by
\[
h(x)=(2\pi)^{-d/2}\bigl(1+(\|x-v_0\|+1)^2\bigr)
e^{-(\|x-v_0\|-1)_+^2/2}.
\]
This function is integrable and does not depend on $v$. Since
$0\leq f\leq1$, the integrands $f(x)\varphi(x-v)$ and their first two
derivatives in $v$ are dominated by $h$. By dominated convergence, we
may differentiate twice under the integral, and the resulting
derivatives are continuous in $v$. Substituting $x=v+z$ back gives the
stated expressions for $\nabla \psi(v)$ and $\nabla^2\psi(v)$.

For the bounds, fix a unit vector $w$ and let $G=\langle w,Z\rangle$,
a standard normal variable. Since $0\leq f\leq1$, the expectation
$\E[f(v+Z)G]$ lies between $-\E G_-$ and $\E G_+$, and
$\E G_+=\E G_-=1/\sqrt{2\pi}$. Thus
$|\langle w,\nabla \psi(v)\rangle|\leq1/\sqrt{2\pi}$. Similarly,
$w^\top\nabla^2\psi(v)w=\E[f(v+Z)(G^2-1)]$ lies between
$-\E(G^2-1)_-$ and $\E(G^2-1)_+$. Since $\E(G^2-1)=0$, both quantities
equal $\tfrac12\E|G^2-1|\leq\tfrac12\sqrt{\E(G^2-1)^2}=\sqrt2/2$.
Taking the supremum over $w$ proves both bounds, since $\nabla^2\psi(v)$ is
symmetric.
\end{proof}

\subsection{Proof of \texorpdfstring{\Cref{lem:efficient-amplification}}{Lemma~\ref{lem:efficient-amplification}}}
\label{app:amplification}

\begin{proof}[Proof of \Cref{lem:efficient-amplification}]
Let $k=|\supp(D)|$, $M=\binom{n+1}{2}$, and $Z\sim D^{n+1}$.
After shuffling, the positions of the copied pair are uniformly distributed.
For each pair of positions, the likelihood ratio relative to $D^{n+1}$ is
the indicator that the entries agree, divided by their common probability.
Therefore, the likelihood ratio of the amplified distribution is
\[
L(Z)=\frac1M\sum_{a<b}\frac{\1\{Z_a=Z_b\}}{D(Z_a)}.
\]
The $\chi^2$ divergence is $\E(L-1)^2$; it therefore remains to calculate
the variance of $L$.
Each summand has mean one and second moment $k$. Moreover, distinct
summands are uncorrelated, since disjoint pairs are independent and, for
pairs sharing one index, the expected product is
\[
\sum_{x\in\supp(D)}\frac{D(x)^3}{D(x)^2}=1.
\]
Since $k\leq d$ and $M=n(n+1)/2>n^2/2$, we obtain
\[
\chi^2\bigl(\Amp(D^n)\Vert D^{n+1}\bigr)
=\E(L-1)^2=\frac{k-1}{M}\leq\frac{2d}{n^2}.
\]
Cauchy--Schwarz gives
\[
\TV\bigl(\Amp(D^n),D^{n+1}\bigr)
=\frac12\E|L-1|\leq\frac12\sqrt{\E(L-1)^2}\leq\frac{\sqrt d}{n}.
\]
Applying \Cref{fact:tv-expectation,fact:chi-square-change-of-measure}
gives the two expectation comparisons, including independent
internal randomness in each experiment.
\end{proof}
\section{Repeated Amplification of the Original Sample}
\label{app:naive-amplification}

In this section we show that the order of amplification and updates in
\Cref{alg:simple-mw} matters. Consider the following variant, which
amplifies the original sample with fresh randomness before every update. Choose
$\mathcal T$ uniformly from $\{0,\ldots,T-1\}$, independently of $S$, and
start from $q_0=e_S$. For $i=1,\ldots,\mathcal T$, independently draw
$(S'_i,Y_i)=\Amp(S)$ and set
\[
q_i=\mathsf U\!\left(q_{i-1},f_i,\frac{f_i(Y_i)+\E_u[f_i]}{2}\right),
\qquad f_i=\dde(S'_i),
\]
with $\mathsf U$ as in \eqref{eq:simple-mw-projection}. Return
$\phat(S)=2\E_{q_{\mathcal T}}[\dde(S)]-\E_u[\dde(S)]$. Each pair
$(S'_i,Y_i)$ is close in distribution to $D^{n+1}$
(\Cref{lem:efficient-amplification}). However, $q_{i-1}$ depends on $S$
and on the earlier pairs, which are all generated from $S$. The pairs
are independent conditional on $S$, but replacing the current pair with
fresh observations does not also replace the sample used to construct
$q_{i-1}$. This prevents the reduction to a complete run on a fresh
sample used in \Cref{lem:simple-mw-recursion}. The following example
shows that the variant itself can fail even when
\Cref{thm:simple-mw} gives a vanishing bound.

\begin{proposition}
\label[proposition]{prop:naive-amplification}
Let $d=2$ and $D(1)=3/4$. For $s\in[2]^n$, let $\dde(s)(x)=\1\{x=1\}$ if
the number of entries of $s$ equal to $1$ is even, and $\dde(s)(x)=0$
otherwise. Suppose that $\eta\to0$ and $\eta T\to\infty$ as $n\to\infty$,
as for the parameters in the proof of \Cref{thm:simple-mw}. Then the mean
squared error of the variant above tends to $9/800$, while the bound in
\Cref{thm:simple-mw} tends to zero.
\end{proposition}

Intuitively, when $S$ has an even number of ones ($k/n\approx 3/4$), an amplified sample $S'_i$ also has an even number of ones if and only if the held-out observation $Y_i$ equals the duplicated entry of $S$. On every update with $f_i\neq 0$, this selection biases $\Prb(Y_i=1)$ from $p=3/4$ to $p^2/(p^2+(1-p)^2)=9/10$, driving the returned estimate toward $9/10$ (and the iterate $q_i(1)$ toward $7/10$) whenever $k$ is even.

\begin{proof}
The bound in \Cref{thm:simple-mw} is
$O(d^{1/4}\sqrt{\log d/n})$, which tends to zero for $d=2$. Fix $S$ with
$k$ entries equal to $1$. If $k$ is odd, then $\dde(S)=0$, and thus
$\phat(S)=0=\E_D[\dde(S)]$. Suppose now that $k$ is even.

We first describe one update. Let $B=1$ if the copied entry in $\Amp(S)$
equals $1$, and $B=0$ otherwise, so that $\Prb(B=1\mid S)=k/n$. The $n+1$
entries then contain $k+B$ ones, and $S'_i$ contains $k+B-\1\{Y_i=1\}$
ones. Thus $f_i\neq0$ exactly when $B=\1\{Y_i=1\}$, and we call such an
update informative. Given $B=1$, we have $Y_i=1$ with probability
$(k+1)/(n+1)$, and given $B=0$, we have $Y_i=2$ with probability
$(n-k+1)/(n+1)$. Therefore an update is informative with probability
\[
\pi_k=\frac{k(k+1)+(n-k)(n-k+1)}{n(n+1)},
\]
and, conditional on this event,
\[
\Prb(Y_i=1\mid f_i\neq0,S)
=p_k=\frac{k(k+1)}{k(k+1)+(n-k)(n-k+1)}.
\]
Conditional on $S$, the updates use independent pairs, and thus the informative
updates occur independently with probability $\pi_k$.

On a non-informative update $f_i=0$, leaving $q_{i-1}$
unchanged. On an informative update $f_i=\1\{\cdot=1\}$ and
$\E_u[f_i]=1/2$, so that the feedback $b_i=(\1\{Y_i=1\}+1/2)/2$ takes values in
$\{1/4,3/4\}$ and has conditional mean $m_k=(p_k+1/2)/2$. Let
$\sigma(z)=1/(1+e^{-z})$ and $\theta_i=\log(q_i(1)/q_i(2))$, so that
$q_i(1)=\sigma(\theta_i)$. The MW step adds $\eta(b_i-\sigma(\theta_{i-1}))$
to $\theta_{i-1}$. The KL projection onto $\widetilde\Delta_2$ maps $q(1)$
to the nearest point of $[1/4,3/4]$; in these coordinates it is therefore the
Euclidean projection onto $I=[-\log3,\log3]$. Since $k/n\to3/4$ in
probability, we have $p_k\to9/10$ and $m_k\to7/10$, and thus with probability
tending to one $\theta^*=\log(m_k/(1-m_k))$ lies in $I$. For such $S$,
the projection does not increase $|\theta-\theta^*|$. On $I$, we have
$\sigma'\geq3/16$, and therefore
$(\theta-\theta^*)(\sigma(\theta)-m_k)\geq\tfrac3{16}(\theta-\theta^*)^2$.
Since $|b_i-\sigma(\theta_{i-1})|\leq1/2$, an informative update satisfies
\[
\E\bigl[(\theta_i-\theta^*)^2\mid\theta_{i-1},S\bigr]
\leq\Bigl(1-\frac{3\eta}{8}\Bigr)(\theta_{i-1}-\theta^*)^2
+\frac{\eta^2}{4}.
\]
The initial point satisfies $q_0(1)=(k/n+1/2)/2\in[1/4,3/4]$, giving
$(\theta_0-\theta^*)^2\leq(2\log3)^2$. After $j$ updates, the number of
informative updates is binomial with parameters $j$ and $\pi_k$, and thus
\[
\E\bigl[(\theta_j-\theta^*)^2\mid S\bigr]
\leq(2\log3)^2\Bigl(1-\frac{3\eta\pi_k}{8}\Bigr)^j+\frac{2\eta}{3}.
\]
Averaging over $\mathcal T$, the first term is at most
$(2\log3)^2/(T(1-(1-3\eta\pi_k/8)))=O(1/(\eta T))$, since $\pi_k\to5/8$.
Both terms tend to zero by our assumptions on $\eta$ and $T$. Since
$\sigma'\leq1/4$, we obtain $\E[(q_{\mathcal T}(1)-m_k)^2\mid S]\to0$.

Conditional on $k$ being even, the estimate is
$\phat(S)=2q_{\mathcal T}(1)-1/2$, and $2m_k-1/2=p_k$.
Thus under this conditioning $\phat(S)\to9/10$ in probability, while
$\E_D[\dde(S)]=D(1)=3/4$. Finally, $k$ is binomial with parameters $n$
and $3/4$, and therefore $\Prb(k\text{ even})=(1+(-1/2)^n)/2\to1/2$. Since all
estimates and answers lie in $[0,1]$, the mean squared error tends to
$\frac12(9/10-3/4)^2=9/800$.
\end{proof}
\section{Computing the Penalized MW Update}
\label{app:mw-projection-computation}

We implement the penalized projection in \eqref{eq:penalized-mw-update}
to the accuracy required by \Cref{cor:penalized-kl-projection,thm:geometric-amplification}. The
projection takes the vectors $e=e_s=(\widehat D_s+u)/2$ and
$q^+=\MW_\eta(q,f,b)$ as input, with
$q\in\widetilde\Delta_d$ and $\eta>0$ as in
\eqref{eq:penalized-mw-update}. We use the penalty $P_s$ and function $H$
from \eqref{eq:mw-empirical-penalty} and \eqref{eq:mw-H-definition}.
At update $i$ of \Cref{alg:general-amplification}, these inputs use
$s=S_{i-1}$, $f=f_{S_{i-1}}$, and $b=b_{S_{i-1}}(Y_i)$. Thus the penalty is centered at the
empirical estimate from the sample that selects the current query.
Let $l=1/(2d)$ and $\Lambda=\log(2d)$, so that $p\geq l\1$ for every
$p\in\widetilde\Delta_d$.
We count arithmetic operations and evaluations of exponentials and logarithms.

\begin{proposition}[Computing the penalized projection]
\label[proposition]{prop:penalized-mw-computation}
Let $\eta>0$, $\lambda\geq0$, and $0<\tau\leq1$. For
\[
A_s(p)=\KL(p\Vert q^+)+\lambda P_s(p),
\qquad P_s(p)=8H(p-e_s)^2,
\]
we can compute $\widehat p\in\widetilde\Delta_d$ satisfying
\begin{equation}
\label{eq:penalized-mw-approx-opt}
\max_{w\in\widetilde\Delta_d}
\bigl\langle\nabla A_s(\widehat p),\widehat p-w\bigr\rangle
\leq\tau
\end{equation}
using $O\!\left(d\log^3\!\left(2+d(1+\eta+\lambda)/\tau\right)\right)$
operations and $O(d)$ space.
In particular, if $\eta$, $\lambda$, and $\tau^{-1}$ are polynomially
bounded in $nd$, the computation takes $O(d\log^3(nd))$ operations.
\end{proposition}

Our algorithm uses three nested binary searches (interval bisections). The objective $\KL(p\Vert q^+)+8\lambda H(p-e)^2$ couples coordinates of $p$ only through the outer square $H(p-e)^2$ and the simplex constraint $\sum_xp(x)=1$. Replacing $16\lambda H(p-e)$ with a scalar parameter $\theta$ linearizes the square into $\theta H(p-e)$; writing $p_\theta$ for the resulting minimizer in \eqref{eq:penalized-mw-projection-path} below, we find the unique fixed point $\theta^*=16\lambda H(p_{\theta^*}-e)$ by an outer binary search (\Cref{alg:penalized-projection,lem:penalized-mw-projection-path}). For a fixed $\theta$, a Lagrange multiplier $\nu$ for $\sum_xp(x)=1$ decouples the $d$ coordinates, and we find $\nu$ by a middle binary search (\Cref{lem:penalized-mw-path-solver}). Finally, for fixed $(\theta,\nu)$, we find each coordinate $p(x)\in[l,l+1/2]$ by an inner binary search on a strictly increasing scalar equation.

By \Cref{cor:penalized-kl-projection}, the approximation condition \eqref{eq:penalized-mw-approx-opt} gives condition~(H4) of \Cref{thm:geometric-amplification} with this $\tau$, contributing $\tau/\eta$ to its error bound~\eqref{eq:geometric-master}; this is the
approximation guarantee used in the proof of
\Cref{thm:mirror-descent-upper-bound}.
To obtain this condition, it suffices to approximate the minimizer $p^*$
of $A_s$ in $\ell_1$. We show this by
bounding $\nabla A_s$ and its Lipschitz constant on
$\widetilde\Delta_d$.
Let $\Gamma=1+\log(1+d)$. We use
$|H(p-e)-H(p'-e)|\leq\Gamma\|p-p'\|_1$, proved in
\eqref{eq:penalized-mw-computation-H-bounds} below.

\begin{lemma}[From $\ell_1$ accuracy to approximate optimality]
\label[lemma]{lem:penalized-mw-gap-accuracy}
Let $p^*$ minimize $A_s$ over $\widetilde\Delta_d$, and let
$C=100d(1+\eta+\lambda)\log^2(2d)=O(d(1+\eta+\lambda)\log^2 d)$.
Every $\widehat p\in\widetilde\Delta_d$ with
$\|\widehat p-p^*\|_1\leq\tau/C$ satisfies \eqref{eq:penalized-mw-approx-opt}.
\end{lemma}

The square in $8\lambda H(p-e)^2$ couples all coordinates of $p$.
Replacing this term with $\theta H(p-e)$ makes the objective a sum of
functions of individual coordinates, allowing us to optimize them
independently once we account for $\sum_xp(x)=1$.
Define
\begin{equation}
\label{eq:penalized-mw-projection-path}
p_\theta=\argmin_{p\in\widetilde\Delta_d}
\bigl\{\KL(p\Vert q^+)+\theta H(p-e)\bigr\},
\qquad \varphi(\theta)=H(p_\theta-e),\quad\theta\geq0.
\end{equation}
Since
$\nabla(8\lambda H^2)=16\lambda H\nabla H$, the optimality conditions
agree when $\theta=16\lambda H(p_\theta-e)=16\lambda\varphi(\theta)$.
The next lemma shows that this equation has a unique solution and bounds
how much $p_\theta$ changes with $\theta$.

\begin{lemma}[Reduction to a scalar equation]
\label[lemma]{lem:penalized-mw-projection-path}
For $\theta,\theta'\geq0$,
\[
\|p_\theta-p_{\theta'}\|_1^2
\leq(\theta'-\theta)
\bigl(\varphi(\theta)-\varphi(\theta')\bigr).
\]
Consequently, $\varphi$ is nonincreasing and
\[
\|p_\theta-p_{\theta'}\|_1\leq\Gamma|\theta-\theta'|,
\qquad
|\varphi(\theta)-\varphi(\theta')|
\leq\Gamma^2|\theta-\theta'|.
\]
If $\lambda>0$, then
$\Psi(\theta)=\theta-16\lambda\varphi(\theta)$ has a unique zero
$\theta^*$ in $[0,\theta_{\max}]$, where
$\theta_{\max}=16\lambda\log(1+d)$, and $p^*=p_{\theta^*}$.
\end{lemma}

For fixed $\theta$, we enforce $\sum_xp(x)=1$ with a Lagrange multiplier
and compute each coordinate by bisection.

\begin{lemma}[Computing $p_\theta$]
\label[lemma]{lem:penalized-mw-path-solver}
For $0\leq\theta\leq\theta_{\max}$ and $0<\varepsilon_1\leq1/2$, we can
compute $\widehat p\in\widetilde\Delta_d$ with
$\|\widehat p-p_\theta\|_1\leq\varepsilon_1$ in
$O(d\log(N/\alpha)\log(1/\alpha))$ operations, where
$\alpha=\varepsilon_1/(5d)$ and
$N=\Lambda+\eta+\theta_{\max}\Gamma$.
\end{lemma}

To deduce \Cref{prop:penalized-mw-computation}, we use bisection to solve
for $\theta$. By \Cref{lem:penalized-mw-gap-accuracy}, it suffices to
approximate $p^*$ within $\tau/C$ in $\ell_1$.
Approximating $p_\theta$ to $\ell_1$ error $\varepsilon_1$ introduces at most
$\kappa=16\lambda\Gamma\varepsilon_1$ error in evaluating $\Psi(\theta)$.
The stopping rule below
will bound the final error by
$\varepsilon_1+2\Gamma\kappa=(1+32\lambda\Gamma^2)\varepsilon_1$.
For $\lambda>0$, we therefore choose
\begin{equation}
\varepsilon=\min\left\{\frac12,\frac{\tau}{C}\right\},
\qquad
\varepsilon_1=\frac{\varepsilon}{1+32\lambda\Gamma^2},
\qquad \kappa=16\lambda\Gamma\varepsilon_1.
\label{eq:penalized-projection-tolerances}
\end{equation}
\Cref{alg:penalized-projection} uses the approximation to $p_\theta$ from
\Cref{lem:penalized-mw-path-solver} to bisect $\Psi$. We prove its
guarantee next, followed by the auxiliary lemmas.

\begin{algorithm}[H]
\caption{Computing the penalized KL projection}
\label{alg:penalized-projection}
\small
\begin{algorithmic}[1]
\Statex \textbf{Input} $q^+$, $e$, $\eta>0$, $\lambda\geq0$, and $0<\tau\leq1$
\If{$\lambda=0$}
    \State \Return $p(x)=\max\{l,cq^+(x)\}$, where $c$ makes $\sum_xp(x)=1$
\EndIf
\State Set $\varepsilon_1$ and $\kappa$ as in \eqref{eq:penalized-projection-tolerances}
\State $(a,b)\gets(0,16\lambda\log(1+d))$
\Statex
\While{$b-a>4\kappa$}
    \State $\theta\gets(a+b)/2$
    \State Compute $\widehat p\in\widetilde\Delta_d$ with $\|\widehat p-p_\theta\|_1\leq\varepsilon_1$
    \State $\widehat\Psi\gets\theta-16\lambda H(\widehat p-e)$
    \If{$|\widehat\Psi|\leq\kappa$}
        \State \Return $\widehat p$
    \ElsIf{$\widehat\Psi>\kappa$}
        \State $b\gets\theta$
    \Else
        \State $a\gets\theta$
    \EndIf
\EndWhile
\State $\theta\gets(a+b)/2$
\State \Return $\widehat p\in\widetilde\Delta_d$ with $\|\widehat p-p_\theta\|_1\leq\varepsilon_1$
\end{algorithmic}
\end{algorithm}

\begin{proof}[Proof of \Cref{prop:penalized-mw-computation}]
If $\lambda=0$, \Cref{fact:kl-projection-computation} computes the exact KL
projection in $O(d)$ operations, which satisfies \eqref{eq:penalized-mw-approx-opt} with $\tau=0$.

Suppose $\lambda>0$. At each midpoint in
\Cref{alg:penalized-projection}, the approximation to $p_\theta$ satisfies
\[
|\widehat\Psi-\Psi(\theta)|
\leq16\lambda\Gamma\|\widehat p-p_\theta\|_1
\leq\kappa.
\]
Thus whenever $|\widehat\Psi|>\kappa$, its sign agrees with that of
$\Psi(\theta)$, and the retained interval contains $\theta^*$.
If $|\widehat\Psi|\leq\kappa$, then $|\Psi(\theta)|\leq2\kappa$;
since $\Psi$ increases at rate at least one, we have
$|\theta-\theta^*|\leq2\kappa$. This also holds when the algorithm
returns the midpoint of an interval of width at most $4\kappa$.
When the algorithm stops, \Cref{lem:penalized-mw-projection-path} implies
\[
\|\widehat p-p^*\|_1
\leq\varepsilon_1+2\Gamma\kappa
=\varepsilon.
\]
\Cref{lem:penalized-mw-gap-accuracy} then implies \eqref{eq:penalized-mw-approx-opt}.

The three bisections choose $\theta$, enforce $\sum_xp(x)=1$, and
solve for each coordinate of $p$. The last two are described in the proof
of \Cref{lem:penalized-mw-path-solver}. Each takes
$O(\log(2+d(1+\eta+\lambda)/\tau))$ steps.
Indeed, $C$, $\theta_{\max}$, and $N$ are polynomially bounded in
$d(1+\eta+\lambda)$; both $1/\varepsilon$ and
$1/\varepsilon_1$ are polynomially bounded in
$d(1+\eta+\lambda)/\tau$; and
$\theta_{\max}/\kappa=\log(1+d)/(\Gamma\varepsilon_1)$.
Each innermost bisection step costs $O(d)$ operations over all coordinates.
We store $O(d)$ coordinates and the endpoints of three intervals, giving
$O(d)$ space.
\end{proof}

\subsection{Proofs of the Auxiliary Lemmas}

\begin{proof}[Proof of \Cref{lem:penalized-mw-gap-accuracy}]
For $h(t)=|t|\log(1+2d|t|)$ and $t\in[-1/2,1/2]$, we have
\[
h'(t)=\sign(t)
\left(\log(1+2d|t|)+\frac{2d|t|}{1+2d|t|}\right),
\qquad |h'(t)|\leq\Gamma.
\]
Moreover, $h'$ is $4d$-Lipschitz, since for $t\neq0$,
\[
h''(t)=\frac{2d}{1+2d|t|}
+\frac{2d}{(1+2d|t|)^2}\leq4d.
\]
For $p,p',e\in\widetilde\Delta_d$, these estimates imply
\begin{equation}
\label{eq:penalized-mw-computation-H-bounds}
H(p-e)\leq\log(1+d),\qquad
|H(p-e)-H(p'-e)|\leq\Gamma\|p-p'\|_1.
\end{equation}
Here $|p(x)-e(x)|\leq1/2$ and $\|p-e\|_1\leq1$.
The MW step also ensures
$e^{-\eta}/(2d)\leq q^+(x)\leq1$.

The constant vector $\1$ in $\nabla A_s$ has zero inner product
with differences of distributions; we therefore let
$g(p)=\log(p/q^+)+\lambda\nabla P_s(p)$. By
\eqref{eq:penalized-mw-computation-H-bounds} and
$\nabla P_s(p)_x=16H(p-e)h'(p(x)-e(x))$,
\[
\begin{aligned}
\|g(p)\|_\infty&\leq \Lambda+\eta+16\lambda\Gamma\log(1+d),\\
\|g(p)-g(p')\|_\infty&\leq \bigl(2d+16\lambda(\Gamma^2+4d\log(1+d))\bigr)\|p-p'\|_1.
\end{aligned}
\]
For the second bound, we split the difference of the products defining
$\nabla P_s$ and apply the bounds above, along with the fact that $\log$
is $2d$-Lipschitz on $[l,1]$.
Since $d\geq2$ and $\Gamma\leq2\log(2d)$, the sum of these two constants is at most $C=100d(1+\eta+\lambda)\log^2(2d)$.
Optimality of $p^*$ gives
$\langle g(p^*),p^*-w\rangle\leq0$ for all
$w\in\widetilde\Delta_d$. Since this set has $\ell_1$ diameter one,
\[
\begin{aligned}
\langle g(\widehat p),\widehat p-w\rangle
&=\langle g(\widehat p)-g(p^*),\widehat p-w\rangle
+\langle g(p^*),\widehat p-p^*\rangle
+\langle g(p^*),p^*-w\rangle\\
&\leq C\|\widehat p-p^*\|_1.
\end{aligned}
\]
Taking the maximum over $w$ proves the lemma.
\end{proof}

\begin{proof}[Proof of \Cref{lem:penalized-mw-projection-path}]
Applying the optimality condition for $p_\theta$ at $p_{\theta'}$ and the
condition for $p_{\theta'}$ at $p_\theta$, adding the two inequalities,
and using convexity of $H$, we obtain
\[
\begin{aligned}
\KL(p_\theta\Vert p_{\theta'})+
\KL(p_{\theta'}\Vert p_\theta)
&=\langle\log p_\theta-\log p_{\theta'},
p_\theta-p_{\theta'}\rangle\\
&\leq(\theta'-\theta)
\bigl(\varphi(\theta)-\varphi(\theta')\bigr).
\end{aligned}
\]
Pinsker's inequality~\cite[Lemma~11.6.1]{CoverThomas2006} bounds the left side below
by $\|p_\theta-p_{\theta'}\|_1^2$. This proves the first claim, and combining
it with \eqref{eq:penalized-mw-computation-H-bounds} gives
the two Lipschitz bounds.

For $\theta'>\theta$, monotonicity gives
$\Psi(\theta')-\Psi(\theta)\geq\theta'-\theta$.
Also $\Psi(0)\leq0$ and $\Psi(\theta_{\max})\geq0$, since
$0\leq\varphi\leq\log(1+d)$. Continuity therefore ensures a unique
zero. At that zero, the gradients of
$\KL(p\Vert q^+)+\theta^*H(p-e)$ and $A_s(p)$ agree at
$p_{\theta^*}$, and thus this point also satisfies the optimality condition
for $A_s$ over $\widetilde\Delta_d$. Thus $p^*=p_{\theta^*}$.
\end{proof}

\begin{proof}[Proof of \Cref{lem:penalized-mw-path-solver}]
Let $r=l+1/2$, so that
$\widetilde\Delta_d=\{p\in[l,r]^d\mid\sum_xp(x)=1\}$.
For $t\in[l,r]$, let
\[
a_x(t)=\log\frac{t}{q^+(x)}+\theta h'(t-e(x)).
\]
This function is increasing with slope at least one, and
$|a_x(t)|\leq N$. For a scalar $\nu$, let $\xi_x(\nu)$ solve
$a_x(t)+\nu=0$ in $[l,r]$, setting $\xi_x(\nu)=l$ if
$a_x(l)+\nu\geq0$, and $\xi_x(\nu)=r$ if
$a_x(r)+\nu\leq0$. Each $\xi_x$ is nonincreasing and $1$-Lipschitz.
Writing $\xi(\nu)=(\xi_x(\nu))_{x\in[d]}$, the optimality conditions for $p_\theta$ say that
$p_\theta=\xi(\nu^*)$, where
\[
\mathcal S(\nu)=\sum_x\xi_x(\nu),\qquad
\mathcal S(\nu^*)=1.
\]
A root lies in $[-N,N]$, since
$\mathcal S(-N)=dr\geq1$ and $\mathcal S(N)=dl=1/2$.
Furthermore, $\mathcal S$ is $d$-Lipschitz and, since all coordinates
move in one direction,
\begin{equation}
\label{eq:penalized-mw-mass-error}
\|\xi(\nu)-p_\theta\|_1=|\mathcal S(\nu)-1|.
\end{equation}

To find a multiplier for which the coordinates sum to one, we bisect
$\nu$ on $[-N,N]$. At each midpoint, we compute
$\widehat\xi_x\in[l,r]$ to error at most $\alpha$ by bisection of
$a_x(t)+\nu$. Writing $\widehat{\mathcal S}=\sum_x\widehat\xi_x$,
the error in this sum is at most $d\alpha$. If
$\widehat{\mathcal S}>1+d\alpha$, we increase $\nu$; if
$\widehat{\mathcal S}<1-d\alpha$, we decrease it; otherwise we stop. We also stop when the bracket has width at most $\alpha$.
At either stopping condition,
\[
|\mathcal S(\nu)-1|\leq2d\alpha,
\qquad |\widehat{\mathcal S}-1|\leq2d\alpha.
\]
We restore total mass one while retaining $\widehat p(x)\geq l$ by returning
\[
\widehat p=l\1+
\frac{\widehat\xi-l\1}{2\widehat{\mathcal S}-1}.
\]
The denominator is positive because $2d\alpha\leq1/5$.
This point belongs to $\widetilde\Delta_d$, and
$\|\widehat p-\widehat\xi\|_1=|\widehat{\mathcal S}-1|$.
Using \eqref{eq:penalized-mw-mass-error}, we obtain
\[
\|\widehat p-p_\theta\|_1
\leq2d\alpha+d\alpha+2d\alpha
=\varepsilon_1.
\]
There are $O(\log(N/\alpha))$ multiplier iterations, each requiring
$d$ coordinate bisections of $O(\log(1/\alpha))$ steps.
\end{proof}
\section{Proofs for Gaussian Distributions}
\label{app:gaussian-proofs}

In this appendix we calculate the $\chi^2$ divergence of the Gaussian amplifier (\Cref{lem:gaussian-amplification}), establish existence, uniqueness, and measurability of the penalized Gaussian mixture update (\Cref{lem:gaussian-mixture-update}), and bound the fourth moments of the SGD iterates (\Cref{clm:gaussian-sgd-moments}).

\subsection{Gaussian Amplification}
\label{sec:gaussian-amplifier-proof}

\begin{proof}[Proof of \Cref{lem:gaussian-amplification}]
The empirical mean has distribution $\mathcal N(\mu,I_d/n)$ and is
independent of the newly generated centered residuals. For each of the
$d$ coordinates, the covariance matrix of $(W_1,\ldots,W_{n+1})$ is
\[
I_{n+1}+\frac{1}{n(n+1)}\1\1^\top.
\]
The eigenvalue in the direction of $\1$ is $1+1/n$, and every
orthogonal direction has eigenvalue $1$. After an orthogonal change of
coordinates, $\Amp(D_\mu^n)$ and $D_\mu^{n+1}$ are products
of one-dimensional Gaussians.
Their means agree, and their variances differ in exactly $d$ coordinates,
with variance $1+1/n$ for the amplified samples and $1$ for independent samples.
Applying the Gaussian divergence formula
(\Cref{fact:gaussian-chi-square}) with $v=1+1/n$ to these coordinates
proves \eqref{eq:gaussian-amplification-chi-square}.
For $n\geq2$, $-\log(1-n^{-2})\leq4/(3n^2)$. If $x=d/n^2\leq1$,
convexity of the exponential implies
\[
(1-n^{-2})^{-d/2}-1
\leq e^{2x/3}-1\leq(e^{2/3}-1)x\leq x\mper
\]
Computing the mean and the $n+1$ outputs takes $O(nd)$ operations.
\end{proof}

\subsection{Existence of the Mixture Update}
\label{app:gaussian-mixture-existence}

\begin{proof}[Completion of the proof of \Cref{lem:gaussian-mixture-update}]
We justify the first variation and the expansion of relative entropy in the
proof of \Cref{lem:gaussian-mixture-update}, and prove existence, uniqueness,
and measurability of the update (equipping $\mathcal P_4$ with the Borel $\sigma$-algebra of weak convergence). Every Gaussian mixture has a density bounded
above by $(2\pi)^{-d/2}$ and below by
$c\exp(-C\|x\|_2^2)$ for some $c,C>0$; the lower bound follows by
restricting its mixing measure to a ball of positive mass.
Since $f_s$ is bounded, $z/G_\nu$ is bounded above and away from zero. Thus
the logarithms of the ratios of these densities grow at most quadratically in
$\|x\|_2$, and therefore all relative entropies in the expansion are finite, since every
mixture with mixing measure in $\mathcal P_4$ has a finite second moment.
Along the segment from $\nu^+$ to $\rho$, for
$0\leq\varepsilon\leq1/2$, the mixture density is at least
$G_{\nu^+}/2$. Its log ratio with $z$ therefore has at most quadratic
growth, uniformly in this range of $\varepsilon$. Both $G_\rho$ and
$G_{\nu^+}$ have finite second moments, and thus dominated convergence
justifies the first variation.

Up to a term independent of $\rho$, the objective in
\eqref{eq:gaussian-mixture-update} is
\begin{equation}
\KL(G_\rho\Vert G_\nu)
+\eta\bigl(\E_{G_\nu}[f_s]-f_s(y)\bigr)\E_{G_\rho}[f_s]
+\lambda P_s(\rho).
\label{eq:gaussian-mixture-objective}
\end{equation}
We consider this objective on all probability measures on $\mathbb R^d$
with the weak topology, taking value $+\infty$ when $P_s(\rho)=\infty$.
The function $v\mapsto\E_{D_v}[f_s]$ is bounded and continuous, since
Gaussian densities are continuous in total variation as their means
vary; thus its integral against $\rho$ is continuous under weak
convergence. Relative entropy and $P_s$ are lower semicontinuous, and
the bounded query term and quartic penalty make sublevel sets compact.
The objective is finite
at $\rho=\delta_{\widehat\mu_s}$ by the density bounds above, and therefore
the minimum is achieved and finite. The Fourier transform of a
Gaussian never vanishes, and thus Gaussian convolution is injective. Strict
convexity of relative entropy in its first argument therefore implies
uniqueness of the mixing measure.

Finally, \eqref{eq:gaussian-mixture-objective} is jointly Borel
measurable in $(s,\nu,y,\rho)$. This follows from joint measurability of
$(s,x)\mapsto f_s(x)$, continuity of Gaussian convolution under weak
convergence, and measurability of integrals against a varying probability
measure. The domain $\mathcal P_4$ is a countable union of the weakly
compact sets $\{\rho\mid\int\|v\|_2^4\,\rho(dv)\leq m\}$,
$m\in\N$. For fixed $(s,\nu,y)$, the objective is finite and lower
semicontinuous in $\rho$ on this domain and achieves its minimum.
These are the hypotheses of the
measurable minimum theorem in \cite[Corollary~1]{BrownPurves1973},
which makes the unique minimizer a measurable function of $(s,\nu,y)$.
\end{proof}

\subsection{Fourth Moments of the Iterates}
\label{app:gaussian-sgd-moments}

\begin{proof}[Proof of \Cref{clm:gaussian-sgd-moments}]
We first bound the probability. Let
$W=(n+1)\|\bar c-\mu\|^2$. Since
$\sqrt{n+1}(\bar c-\mu)\sim\mathcal N(0,I_d)$, $W$ has the $\chi^2$
distribution with $d$ degrees of freedom, and
$\E e^{W/3}=(1-2/3)^{-d/2}=3^{d/2}$. By the choice of $\kappa$ and Markov's
inequality,
\[
\Prb(\|\bar c-\mu\|>\kappa)
=\Prb(W>2d+6\log n)
\leq3^{d/2}e^{-(2d+6\log n)/3}\leq n^{-2},
\]
since $\tfrac12\log3<\tfrac23$.

For the moments, both iterates are within $\kappa$ of a Gaussian
center. By the iterate ball bound \eqref{eq:gaussian-sgd-ball}, $v$ lies within $\kappa$ of
$\widehat\mu_{\bar S}$, and $\widehat\mu_{\bar S}-\mu\sim\mathcal N(0,I_d/n)$. By the
definition of the update, $V$ lies within $\kappa$ of $\bar c$, and
$\bar c-\mu\sim\mathcal N(0,I_d/(n+1))$. For
$X\sim\mathcal N(0,\sigma^2I_d)$, we have
$\E\|X\|^4=d(d+2)\sigma^4$. Using
$(a+b)^4\leq8a^4+8b^4$, $\kappa^4\leq4R^2$, and
$d(d+2)\leq3d^2\leq3n^2R^2$, we obtain
\[
\E\|v-\mu\|^4\leq8\kappa^4+\frac{8d(d+2)}{n^2}\leq56R^2,
\]
and the same bound holds for $\E\|V-\mu\|^4$.
\end{proof}
\section{Proofs for the Lower Bounds}
\label{app:lower-bound-proofs}

In this appendix we prove the small-sample and confidence lower bounds (\Cref{lem:distinct-coordinates,lem:small-sample-lower-bound}), the random code-size lemma (\Cref{lem:random-code-window}), and the large-sample lower bound (\Cref{lem:large-sample-lower-bound}).

\subsection{Small Samples}

\begin{proof}[Proof of \Cref{lem:distinct-coordinates}]
Let $M^{(t)}$ be the value of $M$ after the $t$-th draw is replaced by an
independent copy. Then $|M-M^{(t)}|\leq1$, and $M\neq M^{(t)}$ only if the
old or the new value of the $t$-th draw appears among the other $n-1$
draws, which has probability at most $2(n-1)/m$. The Efron--Stein
inequality~\cite{EfronStein1981,BoucheronLugosiMassart2013} therefore gives
\[
\Var(M)
\leq\frac12\sum_{t=1}^n\Prb(M\neq M^{(t)})
\leq\frac{n(n-1)}m\mper
\]
By Chebyshev's inequality, $|M-\E M|<2n/\sqrt m$ with probability at least $3/4$. This open interval
contains at most $1+4n/\sqrt m\leq 5\max\{1,n/\sqrt m\}$ integers in $[n]$, and therefore some integer
$j\in[n]$ in the interval has probability at least $(3/4)/(5\max\{1,n/\sqrt m\})=(3/20)\min\{1,\sqrt m/n\}$.
\end{proof}

\begin{proof}[Proof of \Cref{lem:small-sample-lower-bound}]
Let $m=\lfloor d/2\rfloor$. Since $d\geq16n$ and $16n$ is even, we have $m\geq8n$ (that is, $n\leq m/8$) and $m\geq d/3$.
We view inner products and linear independence over $\mathbb F_2$.
Draw independent uniform rows $A_1,\ldots,A_m\in\mathbb F_2^{j+1}$ independently of $\theta\sim\mathrm{Unif}(\mathbb F_2^{j+1})$, and then draw $S\sim D_\theta^n$, where $j\in[n]$ maximizes $\Prb(M=j)$.
We call $A_i$ observed if $i$ appears in $S$, so that $M$ counts
the observed rows, and the distribution of the sampled coordinates in $S$ does not depend on $\theta$ or $A$. Conditional on $M=j$ and the set of observed
coordinates, the $j$ observed rows are independent and uniform on
$\mathbb F_2^{j+1}$. Exposing them one by one, the $r$-th row avoids the
span of the previous rows with probability at least $1-2^{r-1-(j+1)}$,
and therefore the $j$ observed rows are linearly independent with
probability
\[
\prod_{r=1}^j\left(1-2^{r-j-2}\right)
=\prod_{t=2}^{j+1}\left(1-2^{-t}\right)
>\prod_{t=2}^\infty\left(1-2^{-t}\right)
\geq\frac12.
\]
Conditional on this event, the $m-j$ unobserved rows are still independent and uniform, and thus the number $K$ of unobserved coordinates on which the two solutions $\theta_0,\theta_1$ disagree has distribution $\mathrm{Bin}(m-j,1/2)$. Since $j\leq n\leq m/8$, we have $K\geq(m-j)/2>m/4$ with probability at least $1/2$.

Let $E_A(S)$ be the event (depending only on $A=(A_1,\ldots,A_m)$ and the coordinates in $S$) that $M=j$, the $j$ observed rows have rank $j$, and the two solutions $\theta_0,\theta_1$ disagree on more than $m/4$ unobserved coordinates. By \Cref{lem:distinct-coordinates} and the bounds above, $\E_A[\Prb(E_A(S))]\geq\frac14\Prb(M=j)=\Omega(\min\{1,\sqrt d/n\})$. We fix a deterministic choice of rows $A$ achieving at least this expectation, and embed $[m]\times\{-1,1\}$ into $[d]$ as $[2m]\subseteq[d]$.

Under the uniform prior on $\theta\in\mathbb F_2^{j+1}$, on the event $E_A(S)$, the posterior distribution given $S$ is uniform on $\{\theta_0,\theta_1\}$, and $\E_{D_{\theta_0}}[\dde_A(S)]-\E_{D_{\theta_1}}[\dde_A(S)]=K/m>1/4$. By the triangle inequality, every estimate $\hat y\in\mathbb R$ errs by more than $1/8>1/16$ on at least one of $\theta_0,\theta_1$, and thus on the event $E_A(S)$ (for any realization of the internal coins of $\phat$) we have $|\phat(S)-\E_{D_\theta}[\dde_A(S)]|>1/16$ with probability at least $1/2$. Averaging over $\theta$ and taking the supremum over $D\in\Delta_d$ gives the tail lower bound, and Markov's inequality implies the mean squared error lower bound.
\end{proof}

\subsection{Choosing the Code Size}
\label{sec:ambiguous-code}

\begin{proof}[Proof of \Cref{lem:random-code-window}]
Let $N=2^\ell$ and let $I$ be the chosen index. Write $V_*$ for the word
at index $I$ and $U_1,\ldots,U_N$ for the other words of $\mathcal C$ in
list order. Given $\ell$, the words of $\mathcal C$ are independent and
uniform, and $I$ is independent of them. Thus, given $\ell$ and $I$, the
words $V_*,U_1,\ldots,U_N$ are independent and uniform on $\{-1,1\}^k$.
In particular, $V_*$ is independent of $\ell$. Let $y$ be the
observation, drawn from $p(\cdot\mid V_*)$. For a word $v$, the
likelihood ratio
\[
Z(v)=\frac{p(y\mid v)}{p_0(y)}
\]
compares the likelihood of $v$ with the average likelihood
$p_0(y)=\E_U\,p(y\mid U)$ of a uniformly random word $U$. Let
\[
L_*=Z(V_*),\qquad Z_s=Z(U_s),\qquad W=\sum_{s=1}^NZ_s.
\]
Given $\mathcal C$ and $y$, Bayes' rule with the uniform prior on
indices gives each index posterior probability proportional to $Z$ of
its word, and thus the chosen index has posterior probability $L_*/(L_*+W)$.
The observation depends only on $V_*$; therefore, given $\ell$, $V_*$ and $y$,
the words $U_s$ are still independent and uniform. Thus the ratios $Z_s$
are conditionally independent with conditional mean
$\E_U\,p(y\mid U)/p_0(y)=1$, and $W$ has conditional mean $N$.

The posterior probability of the chosen index is bounded away from zero
and one when $W$ is within constant factors of $L_*$. We show that this
happens with probability $\Omega(1/\sqrt k)$ in three steps. First, with
constant probability, $\log L_*$ is within $O(\sqrt k)$ of $J$, and a
word drawn from the posterior given $y$ has likelihood ratio at most
$L_*$ with probability at least $1/2$. Next, the random exponent
$\ell$ makes $N$ within a factor of $2$ of $L_*$ with probability
$\Omega(1/\sqrt k)$. Finally, $W$ is then within constant factors of $N$
with constant probability.

For the first step, let
\[
\alpha=\E_U\bigl[Z(U)\1\{Z(U)\leq L_*\}\bigr],
\]
where $U$ is uniform on $\{-1,1\}^k$ and independent of $V_*$ and $y$.
Given $y$, the posterior distribution of a uniform word has density $Z$
with respect to the uniform distribution, and $V_*$ is a draw from this
posterior. Therefore $\alpha=G(L_*)$, where $G(t)=\Prb(L_*\leq t\mid y)$ is
the conditional CDF of $L_*$. We show that, with
probability at least $1/3$ over $V_*$ and $y$,
\[
\alpha\geq\frac12
\qquad\text{and}\qquad
|\log L_*-J|\leq3\sqrt k.
\]
The first event has probability at least $1/2$. Indeed, every real
random variable $X$ with CDF $F$ satisfies
$\Prb(F(X)<u)\leq u$ for $u\in[0,1]$. Applying this to $L_*$ given $y$,
whose CDF is $G$, shows that $\alpha=G(L_*)\geq1/2$
with probability at least $1/2$. Since
$Z(U)\1\{Z(U)\leq L_*\}\leq L_*$, we have $\alpha\leq L_*$, and thus this event
also gives $L_*\geq1/2$. For the second event, the pair $(V_*,y)$ has the
distribution of $(V,Y)$ in the definition of $\Lambda$, giving $\log L_*$
the distribution of $\Lambda$. Since $\Var(\Lambda)\leq k$,
Chebyshev's inequality shows that the second event fails with
probability at most $1/9$. Both events therefore hold with probability
at least $1/2-1/9>1/3$.

On both events, the exponent $\ell_*=\max\{0,\lceil\log_2L_*\rceil\}$
satisfies $L_*\leq2^{\ell_*}\leq2L_*$, where the case $\ell_*=0$ uses
$L_*\geq1/2$. Thus $|\ell_*\log2-\log L_*|\leq\log2$, and $\ell_*$
belongs to $\mathcal I$ since $\log2+3\sqrt k\leq6\sqrt k$. The exponent
$\ell$ is uniform on $\mathcal I$ and independent of $(V_*,y)$, and
$|\mathcal I|=O(\sqrt k)$. Therefore $\ell=\ell_*$ with conditional
probability $\Omega(1/\sqrt k)$, and then $L_*\leq N\leq2L_*$.

It remains to show that, in this case, $W$ lies in $[L_*/4,100L_*]$ with
conditional probability at least $1/20$. By Markov's inequality,
$\Prb(W>100L_*)\leq N/(100L_*)\leq1/50$. The mean of $W$ alone does not
bound $W$ from below, since rare large ratios can carry most of it. We
therefore truncate the ratios at $L_*$ and let
$R=\sum_sZ_s\1\{Z_s\leq L_*\}\leq W$. Its summands are independent
and lie in $[0,L_*]$, with mean $\alpha\geq1/2$, giving
\[
\E R=N\alpha\geq\frac{L_*}2,\qquad
\E R^2\leq NL_*\alpha+(N\alpha)^2.
\]
The Paley--Zygmund inequality gives
\[
\Prb\Bigl(R>\frac{N\alpha}2\Bigr)
\geq\frac14\cdot\frac{(N\alpha)^2}{NL_*\alpha+(N\alpha)^2}
=\frac14\cdot\frac{N\alpha}{L_*+N\alpha}
\geq\frac1{12}.
\]
Thus $L_*/4\leq W\leq100L_*$ with conditional probability at least
$1/12-1/50>1/20$. On this event, the chosen index has posterior
probability $L_*/(L_*+W)\in[1/101,4/5]$. Multiplying the probabilities
of the three steps, this happens with probability
$\Omega(1/\sqrt k)$.

Finally, if some index has posterior probability in $[\gamma,1-\gamma]$,
then so does the most likely index. Its probability is at least $\gamma$,
and if it is a different index, its probability is at most one minus that
of the first index. This proves the claim.
\end{proof}

\subsection{Large Samples}

\begin{proof}[Proof of \Cref{lem:large-sample-lower-bound}]
Let $m=\lfloor d/2\rfloor\geq m_0$ and $n\geq m/8$.
We first verify the hypotheses of \Cref{lem:random-code-window} for the
distributions of \Cref{con:fixed-sample-code-lower-bound}, with $V$
uniform on the whole cube and $Y=S\sim D_V^n$, and then fix a separated
code. Let $T_j=N(j,+1)-N(j,-1)$, where $N(j,\sigma)$ counts occurrences of
$(j,\sigma)$ in $S$. Then, averaging over a uniform word $U$ for the
second identity,
\[
D_v^n(S)=(2m\cosh\beta)^{-n}e^{\beta\langle v,T\rangle},\qquad
\E_U D_U^n(S)=(2m\cosh\beta)^{-n}\prod_j\cosh(\beta T_j).
\]
Therefore
\[
\Lambda=\log\frac{D_V^n(S)}{\E_U D_U^n(S)}
=\sum_{j=1}^m\bigl(\beta V_jT_j-\log\cosh(\beta T_j)\bigr).
\]
For fixed $V$, the observations are independent. Replacing one of them
changes the counts $T_j$ by a total of at most $2$, and each map
$t\mapsto\beta V_jt-\log\cosh(\beta t)$ is $2\beta$-Lipschitz. Thus
$\Lambda$ changes by at most $4\beta$, and the Efron--Stein inequality~\cite{EfronStein1981,BoucheronLugosiMassart2013}
gives $\Var(\Lambda\mid V)\leq8n\beta^2$. The
conditional distribution of $\Lambda$ given $V$ does not depend on $V$, as
seen by replacing every observed sign $\sigma$ at coordinate $j$ with
$\sigma V_j$. Therefore
\[
\Var(\Lambda)\leq8n\beta^2=8a^2m\leq m.
\]
For the mean $J=\E\Lambda$, each observation falls on coordinate $j$ with
probability $1/m$ and has sign $V_j$ with probability $(1+\tanh\beta)/2$, and thus
$\E[V_jT_j]=(n/m)\tanh\beta$. Since $\log\cosh\geq0$,
\[
J\leq\beta\sum_{j=1}^m\E[V_jT_j]=n\beta\tanh\beta\leq n\beta^2=a^2m.
\]
Applying \Cref{lem:random-code-window} with $k=m$, the resulting list has
at most $e^{a^2m+O(\sqrt m)}$ words, and with probability
$\Omega(1/\sqrt m)$ its most likely index given $S$ has posterior
probability in $[c,1-c]$.

The list may contain repeated or close words. Two independent uniform
words are within Hamming distance $m/4$ with probability $e^{-\Omega(m)}$
by the Chernoff bound. The list has at most $e^{2a^2m+O(\sqrt m)}$ pairs, and thus
for sufficiently small $a$ and sufficiently large $m\geq m_0$, two of its entries are within Hamming
distance $m/4$ with probability $e^{-\Omega(m)}=o(1/\sqrt m)$.
Subtracting this failure probability and averaging over the lists fixes
a code with pairwise Hamming distances greater than $m/4$ and the stated
posterior property. Its words are
distinct, allowing us to identify indices with words. The code and its size
are fixed before the prior and the sample are drawn.

Suppose that the most likely codeword given $S$ is $v$, and let
$f_v(j,\sigma)=\1\{\sigma=v_j\}$. Let $w$ be another codeword, and let
$d_H(v,w)>m/4$ be the number of coordinates on which $v$ and $w$ differ.
Since $n\geq m/8$, for $a\leq1/4$ we have $\beta=a\sqrt{m/n}\leq1/\sqrt2<1$, and therefore
\[
\E_{D_v}[f_v]-\E_{D_w}[f_v]
=\tanh\beta\,\frac{d_H(v,w)}m
\geq\frac{\tanh\beta}4
\mathrel{=}\Omega\!\left(\sqrt{\frac mn}\right).
\]
On the event that $\pi=\Prb(V=v\mid S)\in[c,1-c]$, let $\mu_S$ be the
posterior average of $\E_{D_V}[f_v]$ over $V\neq v$. By the separation,
$\E_{D_v}[f_v]-\mu_S\mathrel{=}\Omega(\sqrt{m/n})$. The law of total variance, applied
with the indicator of $\{V=v\}$, gives
\[
\Var(\E_{D_V}[f_v]\mid S)
\geq\pi(1-\pi)\bigl(\E_{D_v}[f_v]-\mu_S\bigr)^2
\mathrel{=}\Omega\!\left(\frac mn\right).
\]
This event has probability $\Omega(1/\sqrt m)$. Embedding $[m]\times\{-1,1\}$ into $[d]$ as $[2m]\subseteq[d]$ and averaging over this event via \Cref{cor:bayes-lower-bound} gives worst-case mean squared error $\Omega(\sqrt m/n)=\Omega(\sqrt d/n)$.
\end{proof}

\fi

\phantomsection
\addcontentsline{toc}{section}{References}
\begingroup
\linespread{1}\selectfont
\raggedright
\printbibliography
\endgroup

\clearpage
\section*{AI methodology}
\label{par:ai-methodology}
The research questions addressed in this work were posed by the authors. The key technical ideas presented in this work were discovered by OpenAI Astra. The authors used several frontier models in exploration, presentation and drafting of this paper.

Using Astra, the authors also obtained stronger lower bounds on the mean
squared error: $\Omega(\min\{d,n/\log n\}^{2/3}/n)$ for $n,d\geq2$, and
in particular $\Omega(d^{2/3}/n)$ for $n=\Omega(d\log d)$.
This rules out the natural conjecture that $\sqrt d/n$ is the tight rate,
suggested by the $\sqrt d$ threshold for constant error up to logarithmic
factors.
We chose not to include these results in this version until their proofs
are fully understood and adequately presented.
The authors are responsible for the results, proofs, and references.

\end{document}